\pdfoutput=1

\documentclass[11pt]{article}

\usepackage[final]{acl}

\usepackage{times}
\usepackage{latexsym}
\usepackage{multirow}
\usepackage{colortbl}
\usepackage[T1]{fontenc}

\usepackage[utf8]{inputenc}

\usepackage{microtype}

\usepackage{inconsolata}

\usepackage{graphicx}

\usepackage{amsmath}

\usepackage{enumitem}
\setlist[itemize]{noitemsep, topsep=0pt}

\title{Converging to a Lingua Franca: Evolution of Linguistic Regions and Semantics Alignment in Multilingual Large Language Models}

\author{Hongchuan Zeng$^1$, Senyu Han$^1$, Lu Chen$^{1,2{\dagger}}$, Kai Yu$^{1,2{\dagger}}$ \thanks{$\dagger$Lu Chen and Kai Yu are the corresponding authors.}\\
$^1$X-LANCE Lab, Department of Computer Science and Engineering \\ MoE Key Lab of Artificial Intelligence, SJTU AI Institute \\Shanghai Jiao Tong University, Shanghai, China \\ $^2$Suzhou Laboratory, Suzhou, China \\
         \texttt{\{charlie68, cnlnpjhsy, chenlusz, kai.yu\}@sjtu.edu.cn}\\}

\begin{document}
\maketitle
\begin{abstract}
Large language models (LLMs) have demonstrated remarkable performance, particularly in multilingual contexts. While recent studies suggest that LLMs can transfer skills learned in one language to others, the internal mechanisms behind this ability remain unclear. We observed that the neuron activation patterns of LLMs exhibit similarities when processing the same language, revealing the existence and location of key linguistic regions. Additionally, we found that neuron activation patterns are similar when processing sentences with the same semantic meaning in different languages. This indicates that LLMs map semantically identical inputs from different languages into a "Lingua Franca", a common semantic latent space that allows for consistent processing across languages. This semantic alignment becomes more pronounced with training and increased model size, resulting in a more language-agnostic activation pattern. Moreover, we found that key linguistic neurons are concentrated in the first and last layers of LLMs, becoming denser in the first layers as training progresses. Experiments on BLOOM and LLaMA2 support these findings, highlighting the structural evolution of multilingual LLMs during training and scaling up. This paper provides insights into the internal workings of LLMs, offering a foundation for future improvements in their cross-lingual capabilities. The
codes are available at: \href{https://github.com/X-LANCE/LinguaFranca}{https://github.com/X-LANCE/LinguaFranca}.
\end{abstract}

\section{Introduction}

\begin{figure}[!htb]
    \centering
    \includegraphics[width=0.5\textwidth]{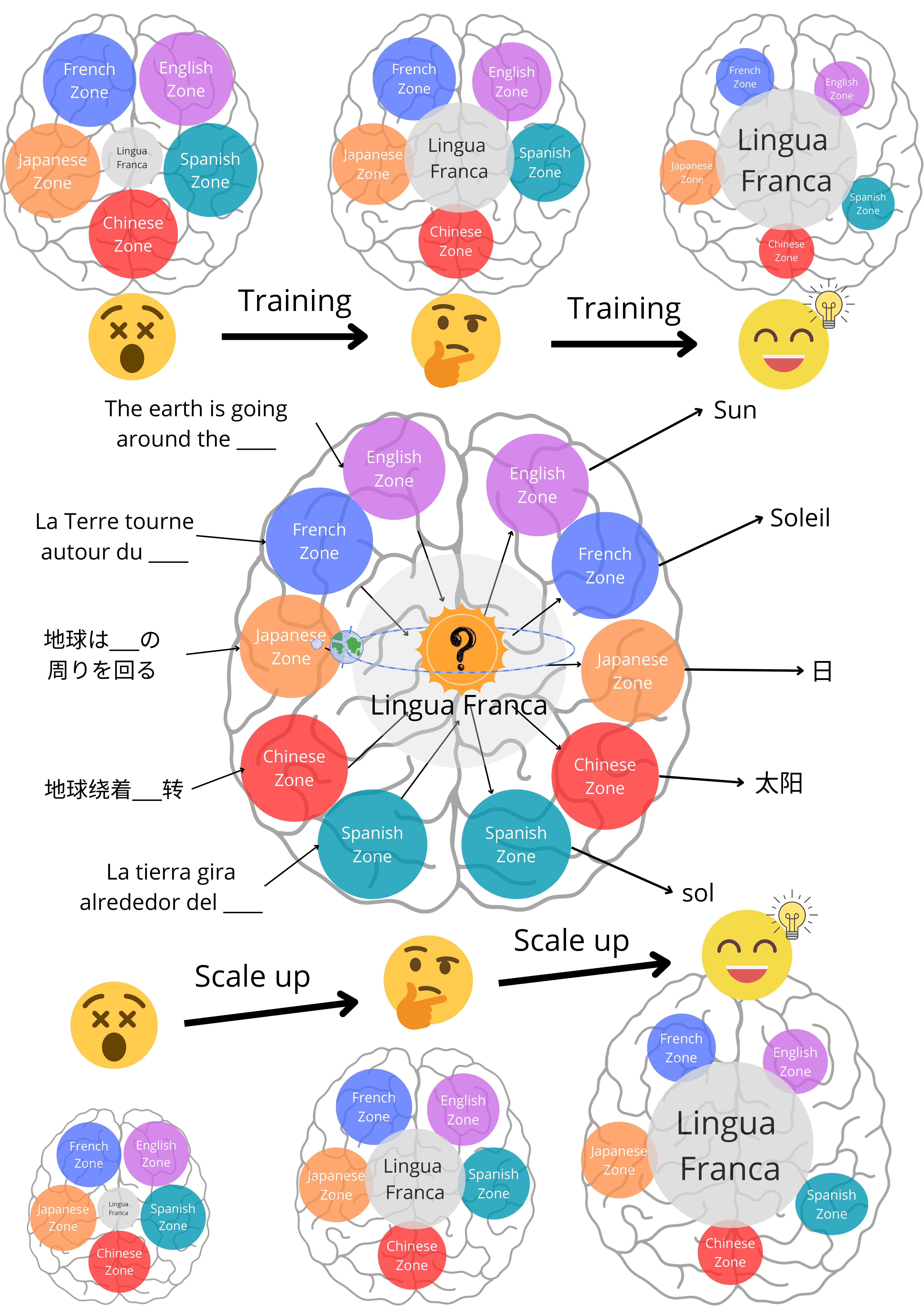}
\caption{LLMs encode inputs into a \textit{"Lingua Franca"}, a latent semantic space representation shared by all languages, and then decode this \textit{"Lingua Franca"} into the target language. As training progresses and the models scale up, LLMs become better at mapping inputs to this common semantic space.}
    \label{soleil}
    \vspace{-\baselineskip} 
\end{figure}

In recent years, large language models (LLMs) have gained significant attention for their remarkable performance. The multilingual capabilities of LLMs are a crucial area of research, especially as AI technology spreads to people with diverse backgrounds and different native languages. Interestingly, recent studies have shown that LLMs can develop cross-lingual abilities, transferring skills learned in one language to others they have not been trained on \citep{chirkova2024zeroshot, pires-etal-2019-multilingual, wu-dredze-2019-beto}. However, the internal mechanisms by which multilingual LLMs function and develop cross-lingual abilities remain an understudied topic. 

In the field of neuroscience, research has shown some interesting findings. First, when polyglots process different languages, their brains' language networks exhibit distinct response patterns \citep{10.1093/cercor/bhae049}. It is believed that different language capacities are stored in different compartments of the human brain \citep{PARADIS19851, article}. Second, while processing the same task in different languages, the human brain exhibits similar activation patterns \citep{XU2021104922}. Third, as individuals gain proficiency in a new language, the activation pattern for that language becomes more similar to those of other languages \citep{nichols2021individual, li2019lexical}. These findings raise the question: Do these phenomena also manifest in LLMs? Our answer is yes. We found that LLMs use different neurons to process inputs in different languages, and map inputs with the same semantic meaning but in different languages into a \textit{"Lingua Franca"}, a common semantic latent space shared by all languages.

First, by examining neuron activation in LLMs, \textit{we observe that neuron activation exhibits similar patterns when processing different inputs in the same language}. \citet{zhang2024unveiling} indicates that certain key parameters in LLMs correspond to linguistic competence, with language-specific parameters existing for different languages. Our research supports this view. By probing the neurons that contribute most to this similarity, we can identify the key linguistic region for a specific language. This key linguistic region consists of neurons responsible for processing specific languages, with each language having its own dedicated region. \textit{When the key linguistic region of a specific language is deactivated, the LLM significantly loses its capacity for the specific language while maintaining its capacity for others.}

Second, we found that \textit{when processing inputs with the same semantic meaning but in different languages, LLM neuron activation shows similar patterns.} This indicates that multilingual LLMs map inputs into a common semantic latent space, allowing them to process information similarly across languages, facilitating cross-lingual ability transfer. We refer to this phenomenon as \textbf{semantic alignment}.

Third, we found that as training progresses, the sizes of key linguistic regions become smaller, and the activation pattern becomes more language-agnostic.
At the same time, semantic alignment becomes more significant. Similarly, as the model scale increases, the activation pattern of neurons become more language-agnostic, but the semantic alignment becomes more pronounced. We defined metrics to facilitate the comparison of linguistic region distinctions and semantic alignment.

By examining the internal structure of LLMs, we found that \textit{the key linguistic region neurons are generally located in the first and last few layers.} As training progresses and model scale increases, these key regions become denser in the first layers. Based on this information, we hypothesize that LLMs encode inputs into a \textit{"Lingua Franca"}, a latent semantic space representation shared by all languages, and then decode this \textit{"Lingua Franca"} into the target language. As training or model size increases, the model can more efficiently map inputs with the same semantic meaning to a common semantic space and then perform reasoning.

The experiments were primarily conducted on BLOOM \citep{workshop2023bloom}, using their released intermediate checkpoints to examine the evolution during the training process. To showcase the extensibility of our results on other models, we equally performed the experiments on LLaMa2 \citep{touvron2023llama}, and we obtained similar results. 

The following is a summary of our observations:

\textbf{Neuron Activation Patterns:}
\begin{itemize}
    \item Neuron activation exhibits similar patterns when processing the same language, revealing the existence and location of key linguistic regions in LLMs. Deactivating these key neurons significantly impairs performance in the corresponding language.
    \item Neuron activation exhibits similar patterns when processing sentences with the same semantic meaning in different languages.
\end{itemize}

\textbf{Dynamics with Training and Scaling Up:}
\begin{itemize}
    \item As the training process progresses, linguistic regions become smaller, while semantic alignment becomes more significant, resulting in a more language-agnostic activation pattern.
    \item As the model’s scale grows, the activation becomes more language-agnostic, and semantic alignment becomes more pronounced.
    \item Important neurons are generally located in the first and last few layers. As training steps increase and model scale grows, key regions become denser in the first layers.
\end{itemize}

\section{Background}

Large language models like GPT-4 \citep{openai2024gpt4}, LLaMA \citep{touvron2023llama}, and OPT \citep{zhang2022opt} have revolutionized natural language processing with their ability to understand and generate nuanced text. Additionally, multilingual large language models such as BLOOM \citep{workshop2023bloom} and XLM-R \citep{conneau2020unsupervised} overcome language barriers by learning universal representations from texts in multiple languages. Multilingual large language models generally incorporate multilingual data in the pretraining stage for better alignment \citep{qin2024multilingual}. These models typically use the transformer architecture \citep{vaswani2023attention} and are decoder-only, with each layer composed of an attention module and a multilayer perceptron module. 

The attention module is crucial in transformer models, allowing the model to focus on different parts of the input sequence, assigning varying importance to each token. It enables the model to dynamically prioritize important information in the input, enhancing its ability to capture dependencies and relationships, thereby improving performance.

A multilayer perceptron (MLP) consists of fully connected neurons with a nonlinear activation function, organized in at least three layers, and is notable for its ability to distinguish data that is not linearly separable. In BLOOM, the operations can be expressed as follows:
\begin{equation}
\small
\mathbf{z} = \mathbf{W}_1 \mathbf{x} + \mathbf{b}_1, \quad
\mathbf{h} = \sigma(\mathbf{z}), \quad
\mathbf{y} = \mathbf{W}_2 \mathbf{h} + \mathbf{b}_2,
\end{equation}

\noindent where \(\mathbf{x}\) is the input vector, \(\mathbf{W}_i\) are the weight matrices, \(\mathbf{b}_i\) are the bias vectors, \(\mathbf{h}\) is the hidden layer output, \(\mathbf{y}\) is the final output, and \(\sigma(\cdot)\) is an activation function. In BLOOM, the activation is GeLU \citep{hendrycks2023gaussian}.

In LLaMA, the MLP module can be expressed as:
\begin{equation}
\small
  \mathbf{h}_1 = \mathbf{W}_1 \mathbf{x} + \mathbf{b}_1, \quad
  \mathbf{h}_2 = \mathbf{W}_2 \mathbf{x} + \mathbf{b}_2, 
  \end{equation}
      \begin{equation}
      \small
  \text{SwiGLU}(\mathbf{x}) = \text{Swish}(\mathbf{h}_1) \odot \mathbf{h}_2, 
  \end{equation}
    \begin{equation}
    \small
  \mathbf{y} = \mathbf{W}_3 \cdot \text{SwiGLU}(\mathbf{x}) + \mathbf{b}_3, 
  \end{equation}

\noindent where Swish function is defined in \citet{ramachandran2017searching}.

As pointed out by \cite{dai2022knowledge}, MLP weights can store complex syntactical and semantic patterns, which are the building blocks of language. Therefore, \textit{in the following parts of our paper, we refer to neurons of BLOOM as \(\mathbf{h}\), representing the hidden layer output, and neurons of LLaMA as the output of \text{SwiGLU}($\mathbf{x}$).}

\section{Methods}

\subsection{Measuring Activation Similarity of Neurons}

In order to decouple the key linguistic regions and the \textit{"Lingua Franca"}, which is the latent semantic space representation shared by all languages, we use a \textbf{parallel corpus} to activate the neurons of the LLM. We record the neuron activation results (e.g. the hidden layer output \(\mathbf{h}\) of MLP layer in BLOOM or the output of $\text{SwiGLU}(\mathbf{x})$ in LLaMA) as the LLM processes each token. We then average these results across all tokens in each sample. By concatenating the averaged results from all layers, we create an activation vector for each sample and normalize it. Finally, we calculate the cosine similarity between each pair of samples to generate the similarity map.

\textbf{Neuron Activation Extraction:}
\begin{equation}
\small
\overline{\mathbf{h}}_{m}^{s_i} = \frac{1}{T_{s_i}} \sum_{t=1}^{T} \mathbf{h}_{m,t}^{s_i}, 
\vspace{-0.5\baselineskip} 
\end{equation}
where \( \mathbf{h}_{m,t}^{s_i} \) represents the activation of token \( t \) in layer \( m \) for sample \( s_i \), and \( T_{s_i} \) is the number of tokens in the sample \( s_i \). Motivated by Sentence-BERT \cite{reimers2019sentencebertsentenceembeddingsusing}, we apply a mean pooling strategy across all tokens to obtain the representation of a sample sentence.

\textbf{Concatenation of Layer Activations:}
\begin{equation}
\small
\mathbf{a}^{s_i} = \frac{1}{||\left[ \overline{\mathbf{h}}_{1}^{s_i} \; \middle| \; \overline{\mathbf{h}}_{2}^{s_i} \; \middle| \; \cdots \; \middle| \; \overline{\mathbf{h}}_{M}^{s_i} \right]||}\left[ \overline{\mathbf{h}}_{1}^{s_i} \; \middle| \; \overline{\mathbf{h}}_{2}^{s_i} \; \middle| \; \cdots \; \middle| \; \overline{\mathbf{h}}_{M}^{s_i} \right], 
\end{equation}
where \( M \) is the total number of layers.

\textbf{Cosine Similarity Map:}
\begin{equation}
\small
S_{ij} = \text{Similarity}(s_i, s_j) = \frac{\mathbf{a}^{s_i} \cdot \mathbf{a}^{s_j}}{\|\mathbf{a}^{s_i}\| \|\mathbf{a}^{s_j}\|}
= \mathbf{a}^{s_i} \cdot \mathbf{a}^{s_j}, 
\end{equation}
as $\mathbf{a}^{s_i}$ is normalized.

\subsection{Metrics on the Development of Linguistic Regions and Semantic Alignment}

To measure how closely related the activation is to language-specific information, we define Linguistic Regions Development Scores (\textbf{LRDS}). Specifically, LRDS measures the difference between the average similarity of samples in the same language and samples in different languages, with all sample pairs having different semantic meanings. The sample numbers in each language are equal. \textit{A lower LRDS indicates that the activation pattern is more language-agnostic.}
\begin{equation}
\small
    \begin{aligned}
        \text{\textbf{LRDS}} &= Average(S_{ij} \mid \text{lang}(s_i) = \text{lang}(s_j), \\
       &\quad \quad \quad \quad \text{semantics}(s_i) \neq \text{semantics}(s_j)) \\
        &\quad - Average(S_{ij} \mid \text{lang}(s_i) \neq \text{lang}(s_j), \\
        &\quad \quad \quad \quad \text{semantics}(s_i) \neq \text{semantics}(s_j)). 
    \end{aligned}
\end{equation}

The Size of Key Linguistic Regions (\textbf{SKLR}) is the sum of the sizes of the key linguistic regions for all languages, which we will present further in Section \ref{KLR}. This metric evaluates how computationally costly it is to align inputs in different languages to the semantic space.

To measure how closely related the activation is to the semantic meaning of the inputs instead of the language of the inputs, we define Semantic Alignment Development Scores (\textbf{SADS}). Specifically, SADS measures the difference between the average similarity of samples with the same meaning and samples with different meanings, with all sample pairs in different languages. \textit{A higher SADS indicates that the activation pattern is more related to the semantic meaning of the inputs.}

\begin{equation}
\small
    \begin{aligned}
        \text{\textbf{SADS}} &= Average(S_{ij} \mid \text{semantics}(s_i) = \text{semantics}(s_j), \\
       &\quad \quad \quad \quad \quad \quad \quad \quad \text{lang}(s_i) \neq \text{lang}(s_j) ) \\
        & - Average(S_{ij} \mid \text{semantics}(s_i) \neq \text{semantics}(s_j), \\
        &\quad \quad \quad \quad \quad \quad \quad \quad \text{lang}(s_i) \neq \text{lang}(s_j)). 
    \end{aligned}
    \vspace{-0.5\baselineskip} 
\end{equation}

\subsection{Key Linguistic Region Probing}
\label{KLR}

We believe that neurons activated in a similar pattern across different samples in one language are key neurons for that specific language. Therefore, we assign a score to these neurons to evaluate their contribution to the average similarity across the samples in that specific language.

The average similarity $\overline S_l$ of samples in one specific language $l$ can be expressed by :
\begin{equation}
\small
    \begin{aligned}
    \overline S_l &= \frac{2}{n(n-1)} \sum_{i=1}^{n-1} \sum_{j=i+1}^{n} S_{ij} \\
    &=  \frac{2}{n(n-1)} \sum_{i=1}^{n-1} \sum_{j=i+1}^{n} \mathbf{a}^{s_i} \cdot \mathbf{a}^{s_j}\\
    &= \frac{2}{n(n-1)} \sum_{i=1}^{n-1} \sum_{j=i+1}^{n} \sum_{k=1}^{K} \mathbf{a}^{s_i}_{(k)} \cdot \mathbf{a}^{s_j}_{(k)} \\
    &= \frac{2}{n(n-1)}\sum_{k=1}^{K} ( \sum_{i=1}^{n-1} \sum_{j=i+1}^{n} \mathbf{a}^{s_i}_{(k)} \cdot \mathbf{a}^{s_j}_{(k)}), 
    \end{aligned}
\end{equation}

\noindent where $n$ is the total number of sample in language $l$ $\mathbf{a}^{s_i}_{(k)}$ is the activation of neuron $k$ (the $k$ component of activation vector $\mathbf{a}^{s_i}$) for a sample ${s_i}$, and $K$ the total number of neurons (i.e. the length of activation vector).

We may see that, the contribution of neuron $k$ to the average cosine similarity can be quantified by the term in the bracket. We thus define the contribution score $\overline S^{(k)}_l$ of a neuron $k$ to one language $l$ as:
\begin{equation}
\small
\overline S^{(k)}_l = \sum_{i=1}^{n-1} \sum_{j=i+1}^{n} \mathbf{a}^{s_i}_{(k)} \cdot \mathbf{a}^{s_j}_{(k)}.
\end{equation}

Since some neurons are always activated, their contribution to the average similarity is consistently large. Therefore, we need to identify neurons that contribute exceptionally to the average similarity for a specific language. To do this, we calculate the standard score (z-scores, \( z^{(k)}_l \)) of the contribution score of a neuron $k$ across different languages: 
\begin{equation}
\small
z^{(k)}_l = \frac{\overline{S}^{(k)}_l - \mu_k}{\sigma_k}, 
\end{equation}

\noindent where \( \mu_k \) and \( \sigma_k \) are the mean and standard deviation of the contribution scores of the neuron \( k \) across all languages, respectively:
\begin{equation}
\small
\mu_k = \frac{1}{L} \sum_{l=1}^{L} \overline{S}^{(k)}_l, 
\vspace{-0.5\baselineskip} 
\end{equation}

\begin{equation}
\small
\sigma_k = \sqrt{\frac{1}{L} \sum_{l=1}^{L} (\overline{S}^{(k)}_l - \mu_k)^2}. 
\end{equation}
Here, \( L \) is the total number of languages. The value \( z^{(k)}_l \) represents the extent to which the neuron $k$ contribute exceptionally to language $l$. In practice, we set a threshold for the z-scores, and neurons with z-scores higher than this threshold for a specific language are considered part of the key linguistic region for that language. For example, if the threshold is set to 2, the neurons we select have activation scores that are more than 2 standard deviations above their average scores.

\section{Experiments}

We first examine the similarity of activation patterns within an individual model. To illustrate the language-wise and semantic-wise similarity when LLMs process inputs, we plot the similarity map of neuron activation while processing inputs in the same language or with the same semantic meaning.

Next, using the obtained similarity patterns, we locate the key linguistic region for each language. By deactivating the key region of each language respectively, we assess the performance of LLMs, demonstrating the existence and utility of these regions.

We then explore the dynamic changes during training and scaling up. We calculate Linguistic Regions Development Scores (\textbf{LRDS}), Size of Key Linguistic Regions (\textbf{SKLR}), and Semantic Alignment Development Scores (\textbf{SADS}) for different training checkpoints to illustrate the development of linguistic regions and semantic alignment. We perform the same analysis for models of different scales to show the evolution trend.

\subsection{Experimental Setup}

\textbf{Models.} The experiments were primarily conducted using the BLOOM \citep{workshop2023bloom} model family, which features a high percentage of multilingual training data and is well-balanced across various languages. We tested various models within this family, including the 560m, 1.1b, 1.7b, 3B, and 7.1b models, as well as the intermediary checkpoints of BLOOM-7b1. Additionally, we conducted complementary experiments on the LLaMA-2 model family to examine scenarios where multilingual training data is limited.

\textbf{Datasets \& Language Selection.} We used the Bible dataset \citep{christodouloupoulos2015massively}, a perfectly aligned parallel corpus, to activate the LLM. Each sample consists of a verse, and we randomly selected 100 verses, taking translations in different languages from the dataset. To evaluate multilingual perplexity, we employed the XL-Sum dataset \citep{hasan-etal-2021-xl}, following the implementation of \citet{zeng-etal-2024-multilingual-brain}. XL-Sum contains high-quality articles from the BBC covering 45 languages. Our experiments focused on a subset of 9 languages available in both BLOOM and XL-Sum: Arabic (\texttt{ar}), Chinese (\texttt{zh}), English (\texttt{en}), French (\texttt{fr}), Hindi (\texttt{hi}), Indonesian (\texttt{id}), Portuguese (\texttt{pt}), Spanish (\texttt{es}), and Vietnamese (\texttt{vi}).

\textbf{Evaluation.} We evaluated the perplexity of the models separately for each language using XL-Sum. Additionally, we designed a task to assess the cross-lingual reasoning ability of LLMs, employing the widely recognized EleutherAI-eval-harness framework \citep{eval-harness}. For this, we used the XStoryCloze dataset, which consists of a short story typically composed of four sentences and two alternative endings. One ending logically completes the story, while the other does not. The task for the model is to identify the more plausible ending. We prompted the story in various languages and asked the model to choose between the two ending options in English. Prompting the stories and endings in different languages helps us determine whether the model effectively "understands" the stories and maps them into a mutual semantic space, rather than performing the reasoning process in only one language. The prompt format and evaluation details are presented in Appendix \ref{sec:XStoryCloze}. 

\begin{figure}[!htb]
    \centering
    \includegraphics[width=0.45\textwidth]{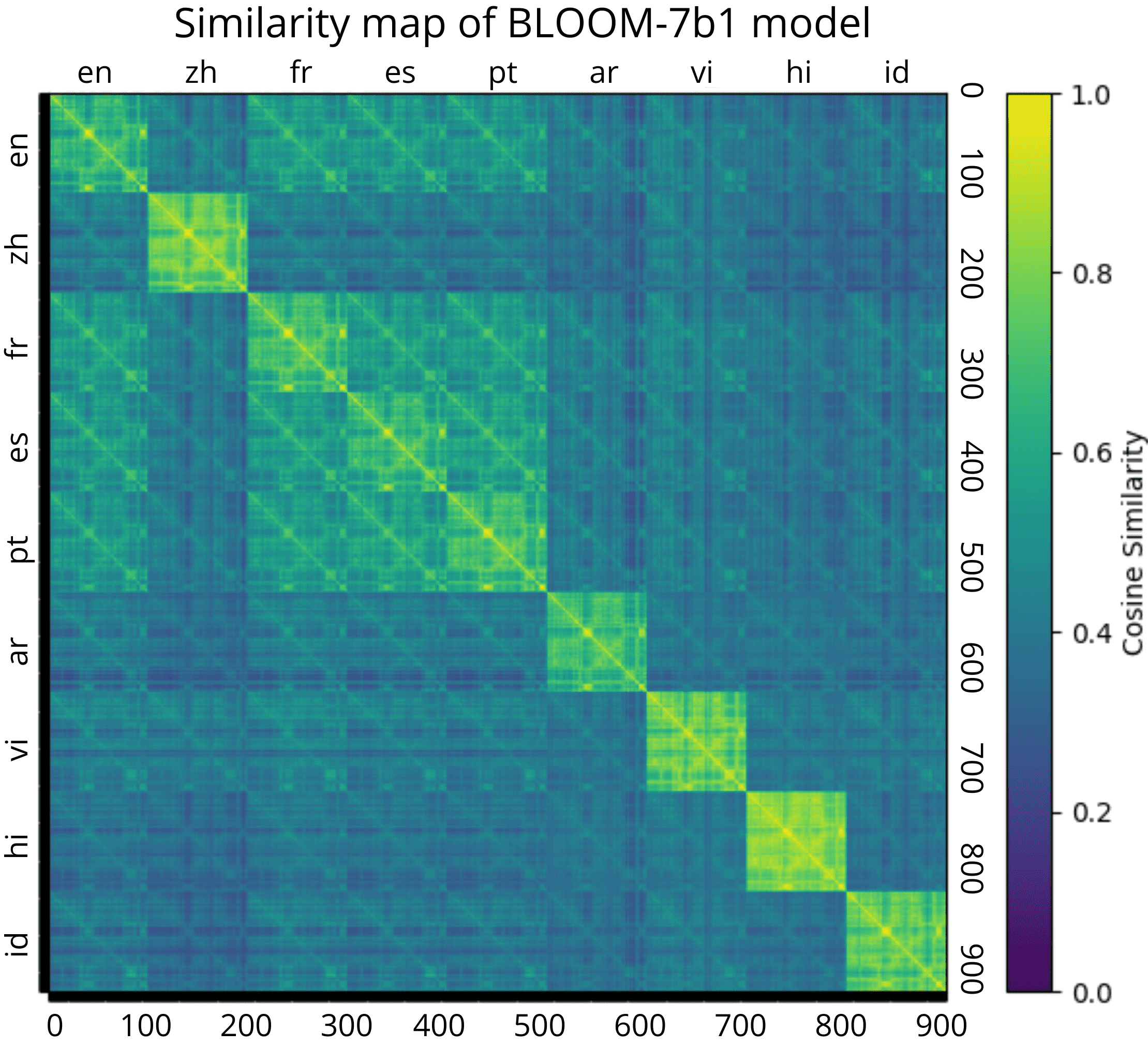}
\caption{Similarity map of the BLOOM-7b1 model. Each block of 100 samples is in the same language. \textit{Samples in the same language form distinct light blocks, and samples with the same semantic meaning form light bands along the diagonal of these blocks.}}
    \label{SimilarityMap7b1}
    \vspace{-\baselineskip}
\end{figure}

\begin{table*}[!htb]
\centering
\small
\begin{tabular}[width=0.8\textwidth]{llllllllllll}
\rowcolor[HTML]{BFBFBF} 
\begin{tabular}[c]{@{}l@{}}Perplexity\\ increase \%↑\end{tabular}                    & \begin{tabular}[c]{@{}l@{}}Full model\\ Perplexity\end{tabular} & \begin{tabular}[c]{@{}l@{}}Random\\ 10\%\end{tabular} & $\times$en                    & $\times$zh                    & $\times$fr                    & $\times$es                   & $\times$pt                   & $\times$ar                    & $\times$vi                    & $\times$hi                    & $\times$id                     \\
en                                                                                   & 13.94                                                           & 12\%                                                  & \cellcolor[HTML]{F8696B}22\%  & \cellcolor[HTML]{FFEA84}6\%   & \cellcolor[HTML]{FFEA84}2\%   & \cellcolor[HTML]{FFE984}2\%  & \cellcolor[HTML]{FFEB84}1\%  & \cellcolor[HTML]{FFEB84}55\%  & \cellcolor[HTML]{FFE784}18\%  & \cellcolor[HTML]{C8DB80}7\%   & \cellcolor[HTML]{F6E883}5\%    \\
zh                                                                                   & 24.01                                                           & 11\%                                                  & \cellcolor[HTML]{A5D17E}3\%   & \cellcolor[HTML]{F8696B}47\%  & \cellcolor[HTML]{FCEA83}2\%   & \cellcolor[HTML]{98CD7E}1\%  & \cellcolor[HTML]{B8D67F}1\%  & \cellcolor[HTML]{63BE7B}47\%  & \cellcolor[HTML]{FFEA84}16\%  & \cellcolor[HTML]{FFEB84}8\%   & \cellcolor[HTML]{C3D980}4\%    \\
fr                                                                                   & 9.62                                                            & 10\%                                                  & \cellcolor[HTML]{FFEA84}5\%   & \cellcolor[HTML]{FFEB84}5\%   & \cellcolor[HTML]{F8696B}20\%  & \cellcolor[HTML]{FFDC82}4\%  & \cellcolor[HTML]{FFE884}2\%  & \cellcolor[HTML]{86C87D}49\%  & \cellcolor[HTML]{A3D07E}14\%  & \cellcolor[HTML]{FFEB84}8\%   & \cellcolor[HTML]{FFEB84}5\%    \\
es                                                                                   & 10.84                                                           & 10\%                                                  & \cellcolor[HTML]{FFEB84}5\%   & \cellcolor[HTML]{B8D67F}5\%   & \cellcolor[HTML]{FFEB84}2\%   & \cellcolor[HTML]{F8696B}17\% & \cellcolor[HTML]{FFE183}3\%  & \cellcolor[HTML]{76C37C}48\%  & \cellcolor[HTML]{63BE7B}14\%  & \cellcolor[HTML]{FDEA83}8\%   & \cellcolor[HTML]{FFEB84}5\%    \\
pt                                                                                   & 11.17                                                           & 11\%                                                  & \cellcolor[HTML]{FFEB84}5\%   & \cellcolor[HTML]{D2DE81}5\%   & \cellcolor[HTML]{FFEB84}2\%   & \cellcolor[HTML]{FEC77E}6\%  & \cellcolor[HTML]{F8696B}27\% & \cellcolor[HTML]{A5D17E}50\%  & \cellcolor[HTML]{CEDD81}15\%  & \cellcolor[HTML]{FFEB84}8\%   & \cellcolor[HTML]{FFEB84}6\%    \\
ar                                                                                   & 14.45                                                           & 12\%                                                  & \cellcolor[HTML]{EEE683}4\%   & \cellcolor[HTML]{D6DF81}5\%   & \cellcolor[HTML]{D5DE81}2\%   & \cellcolor[HTML]{ECE582}2\%  & \cellcolor[HTML]{FAE983}1\%  & \cellcolor[HTML]{F8696B}309\% & \cellcolor[HTML]{FFEB84}16\%  & \cellcolor[HTML]{63BE7B}6\%   & \cellcolor[HTML]{D4DE81}4\%    \\
vi                                                                                   & 10.11                                                           & 12\%                                                  & \cellcolor[HTML]{EFE683}4\%   & \cellcolor[HTML]{FFE884}7\%   & \cellcolor[HTML]{FFEB84}2\%   & \cellcolor[HTML]{B1D47F}2\%  & \cellcolor[HTML]{FFEB84}1\%  & \cellcolor[HTML]{FFE784}64\%  & \cellcolor[HTML]{F8696B}83\%  & \cellcolor[HTML]{9ECF7E}7\%   & \cellcolor[HTML]{FFEB84}6\%    \\
hi                                                                                   & 11.14                                                           & 10\%                                                  & \cellcolor[HTML]{63BE7B}3\%   & \cellcolor[HTML]{63BE7B}4\%   & \cellcolor[HTML]{63BE7B}1\%   & \cellcolor[HTML]{63BE7B}1\%  & \cellcolor[HTML]{63BE7B}1\%  & \cellcolor[HTML]{FED07F}109\% & \cellcolor[HTML]{C9DB80}15\%  & \cellcolor[HTML]{F8696B}220\% & \cellcolor[HTML]{63BE7B}3\%    \\
id                                                                                   & 20.55                                                           & 13\%                                                  & \cellcolor[HTML]{FFE884}5\%   & \cellcolor[HTML]{FFE784}7\%   & \cellcolor[HTML]{DCE182}2\%   & \cellcolor[HTML]{FFEB84}2\%  & \cellcolor[HTML]{F0E683}1\%  & \cellcolor[HTML]{FFDF82}80\%  & \cellcolor[HTML]{FFE884}18\%  & \cellcolor[HTML]{FFE984}12\%  & \cellcolor[HTML]{F8696B}1498\% \\
\cellcolor[HTML]{D9E1F4}\begin{tabular}[c]{@{}l@{}}Key Neuron \\ Number\end{tabular} & \cellcolor[HTML]{D9E1F4}                                        & \cellcolor[HTML]{91AADF}49152                         & \cellcolor[HTML]{CAD6F0}15935 & \cellcolor[HTML]{B0C2E8}31185 & \cellcolor[HTML]{CED9F1}14040 & \cellcolor[HTML]{D9E1F4}7182 & \cellcolor[HTML]{D7DFF4}8865 & \cellcolor[HTML]{96AEE1}46313 & \cellcolor[HTML]{97AFE1}45758 & \cellcolor[HTML]{C0CEED}22285 & \cellcolor[HTML]{CCD7F0}15201  \\
\rowcolor[HTML]{BFBFBF} 
\begin{tabular}[c]{@{}l@{}}Key Neuron \\ Percentage\end{tabular}                     &                                                                 & 10\%                                                  & 3.2\%                         & 6.3\%                         & 2.9\%                         & 1.5\%                        & 1.8\%                        & 9.4\%                         & 9.3\%                         & 4.5\%                         & 3.1\%                         
\end{tabular}
\caption{Percentage increase in perplexity after deactivating key linguistic region neurons for each language of BLOOM-7b1 model. Each column corresponds to the deactivation of the key region for a specific language. The second column shows the results of deactivating a random 10\% of neurons in the LLM. \textit{We can see that the perplexity of the deactivated language (on the diagonal) rises significantly, while the perplexity of other languages remains largely unchanged.}}
    \label{deactivatenNeurons}
    \vspace*{-\baselineskip}
\end{table*}

\subsection{Observation I: Neuron activation exhibits similar pattern when processing different inputs in the same language or with the same semantic meaning.}

To illustrate the language-wise and semantic-wise similarity when LLMs process inputs, we used the BLOOM-7b1 model as an example and plotted the similarity map of all the sample pairs, as shown in Figure \ref{SimilarityMap7b1}. Each block of 100 samples is in the same language, and sentences with the same semantic meaning are placed in the same position across all sample blocks. \textit{Within each language block, the similarity of neuron activation is significantly higher, indicating that LLMs' neuron activation exhibits similar patterns when processing inputs in the same language.}

Additionally, \textit{we observed bright bands on the diagonal of the off-diagonal blocks.} These represent sentences in different languages but with the same semantic meaning. Even for languages like Chinese and Arabic, which use different alphabets and share no common tokens with other languages, the activation patterns are similar when processing semantically identical sentences. This suggests that the LLM "understands" the semantic meaning of sentences, encoding the inputs into a common semantic space shared by all languages, allowing it to process information similarly across languages.

Interestingly, we also observed that linguistically similar languages, such as French, Spanish, and Portuguese, exhibit higher cross-lingual similarity compared to other language pairs. This could be because these languages share more tokens and have similar grammatical syntax, leading the LLM to process them in a similar way.

\begin{figure}[!htb]
    \centering
    \includegraphics[width=0.49\textwidth]{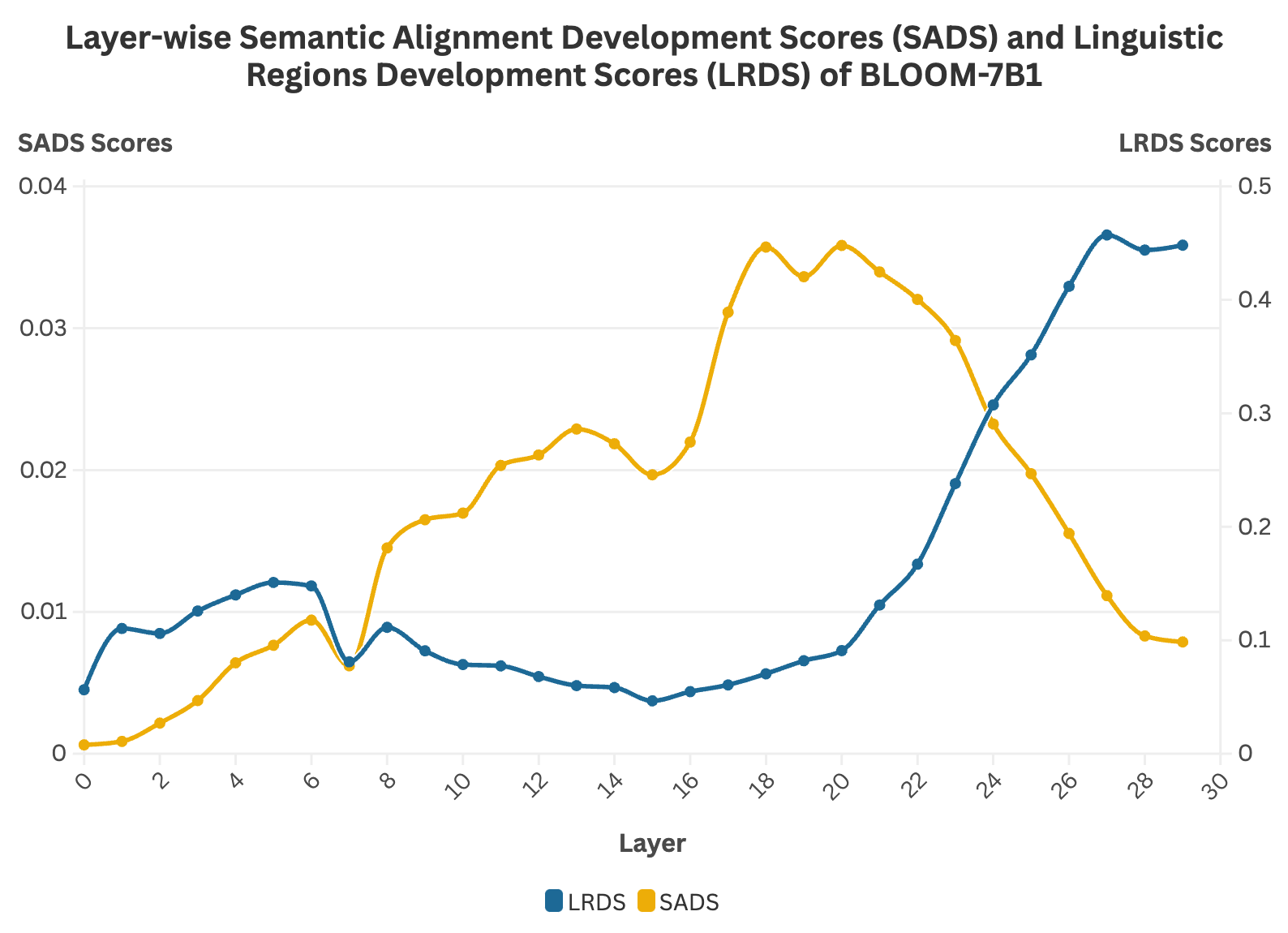}
    \caption{Layer-wise Semantic Alignment Development Scores (SADS) and Linguistic Regions Development Scores (LRDS) of BLOOM-7B1.}
    \label{layerwise}
\end{figure}

We conducted a layer-wise analysis of the BLOOM-7B1 model, calculating both Linguistic Regions Development Scores (LRDS) and Semantic Alignment Development Scores (SADS) for each layer. The results in Figure \ref{layerwise} show that the first and last layers have higher LRDS and lower SADS, indicating a stronger focus on language-specific information with less emphasis on semantic alignment. In contrast, the middle layers have lower LRDS and higher SADS, suggesting a shift towards greater semantic alignment and more language-agnostic processing. \textit{This indicates that the first and last layers are more language-specific and less semantically focused, whereas the middle layers are more semantically oriented and language-agnostic.}

\subsection{Observation II: Similarity of activation patterns allows locating the Key linguistic region of a specific language}

To identify the key linguistic region for each language, we calculated the z-scores of each neuron for each language and selected those neurons with a z-score higher than 2 (i.e., their activation scores for a specific language are more than two standard deviations above the average). These selected neurons contribute exceptionally to the specific language being analyzed. We then deactivated the key neurons for one specific language and measured the perplexity of the model across all languages. The results are shown in Table \ref{deactivatenNeurons}.

\textit{We found that after removing the key region for one language, the perplexity for that language increased significantly}, as shown in the diagonal blocks of the table. However, the perplexity for other languages remained almost the same, as shown in the off-diagonal blocks. This indicates that the neurons we identified are crucial for processing the specific language being researched, forming the key linguistic region for that language.

\subsection{Observation III: As the training process progresses, the linguistic regions become smaller, and the activation patterns become more language-agnostic.}

\begin{figure*}[!htb]
    \centering
    \includegraphics[width=0.95\textwidth]{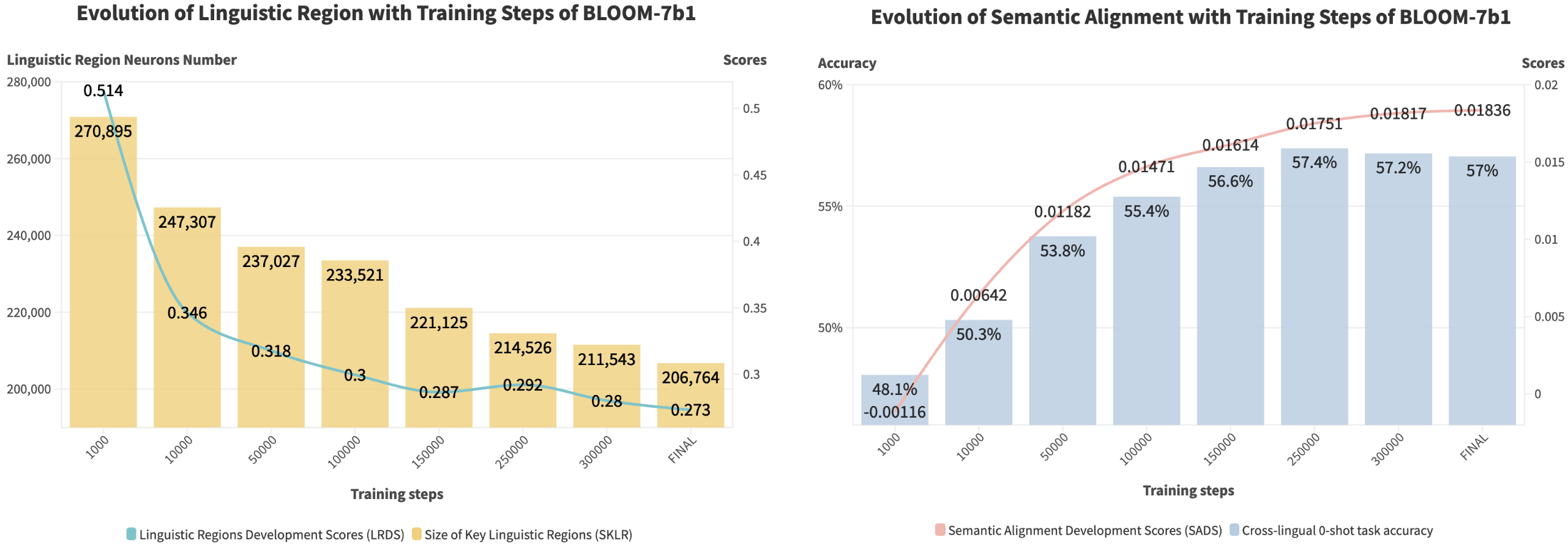}
    \caption{Comparison of the evolution of linguistic regions (left) and semantic alignment (right) with training steps. 
    \textit{As training progresses, the key linguistic regions become smaller, and the neuron activation pattern becomes more language-agnostic. Meanwhile, semantic alignment becomes more pronounced, and the model's cross-lingual reasoning ability improves.}}
    \label{LRDS}
    \vspace*{-\baselineskip}
\end{figure*}

We can now compare the development of linguistic regions throughout the training process. Based on Observation II, we can locate the key linguistic regions for each language, count the number of key neurons, and thus determine the Size of Key Linguistic Regions (\textbf{SKLR}), which is the sum of the sizes of the key linguistic regions for all languages. We can also calculate the Linguistic Regions Development Scores (\textbf{LRDS}) using the similarity map. By applying these metrics to the training checkpoints of BLOOM, we can observe the dynamics of key linguistic region development. The results are shown in Figure \ref{LRDS} (left).

At the beginning of the training process, the size of the key linguistic regions is large, indicating a high number of key neurons. At the same time, the LRDS is also high, signifying that the activation pattern is highly language-specific. However, as training progresses, the size of the key linguistic regions decreases, and the LRDS drops. \textit{This indicates that the activation pattern becomes less related to the specific language and more focused on the semantic meaning of the inputs.} This shift occurs because the model becomes more familiar with the languages and requires less effort to "understand" the sentences and project them into the common semantic space.

\begin{figure}[!htb]
    \centering
    \includegraphics[width=0.5\textwidth]{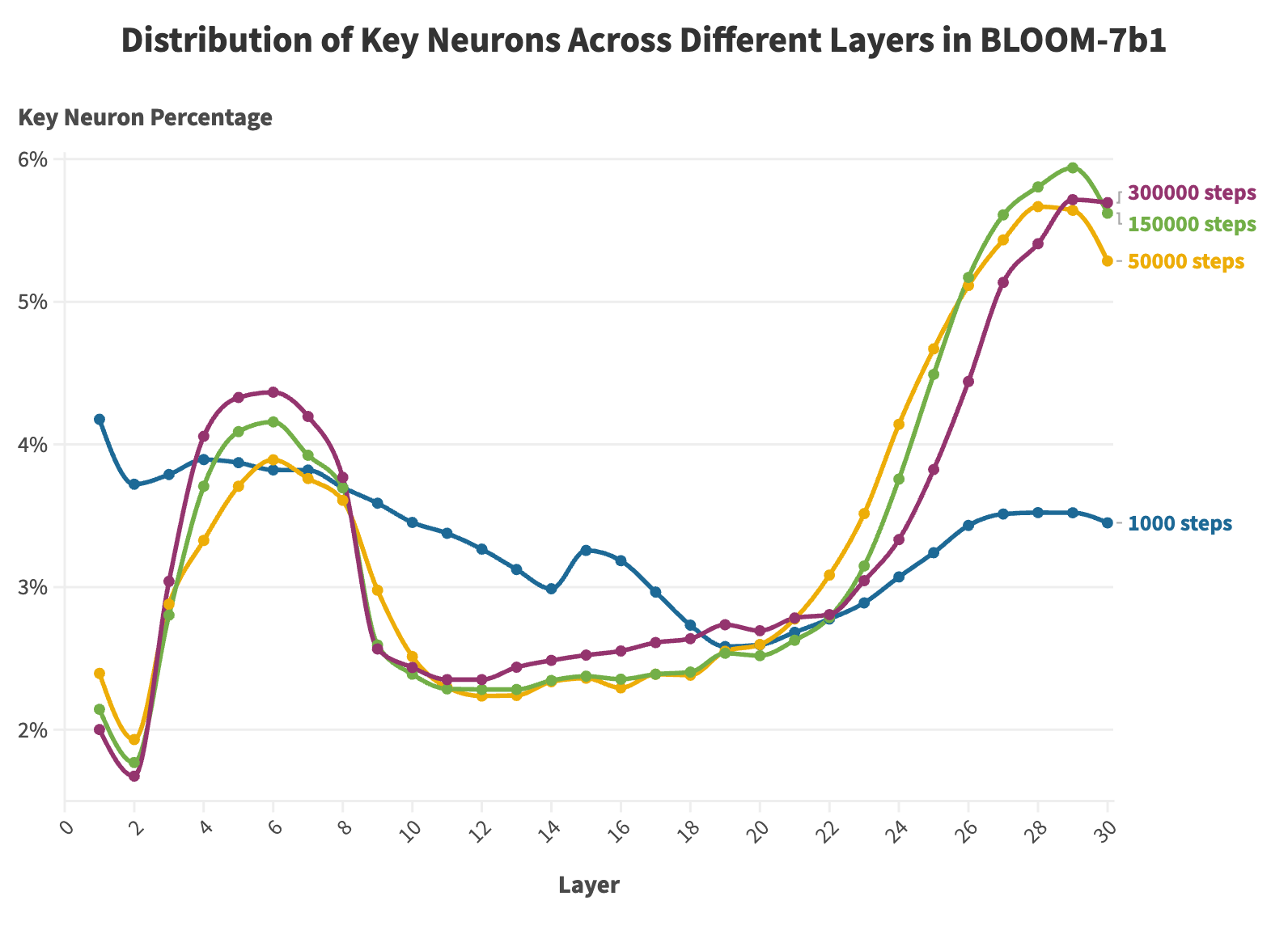}
    \caption{Distribution of Key Neurons Across Different Layers. 
    We may see that as the training steps grows, the key regions become denser in the first layers, facilitating the encoding from inputs to the \textit{"Lingua Franca"}.}
    \label{NeuronsAcrossLayers}
    \vspace{-\baselineskip}
\end{figure}

If we examine the distribution of key neurons, we find that \textit{the key neurons are generally located in the first and last few layers}. As the number of training steps increases, these key regions become denser in the first few layers, as shown in Figure \ref{NeuronsAcrossLayers}. The first few layers are likely related to the encoding process from the source language to the common semantic space, while the latter layers correspond to the decoding process from the latent semantic representation to the target language. As training progresses, the model can more efficiently encode information from different languages into the semantic space and then decode this representation using fewer neurons into the target language.

\subsection{Observation IV: As the training process progresses, the semantic alignment phenomenon becomes more significant.}

To examine the semantic alignment phenomenon throughout the training process, we evaluated the Semantic Alignment Development Scores (\textbf{SADS}) and the cross-lingual zero-shot performance at different checkpoints of BLOOM-7b1. The results are shown in Figure \ref{LRDS} (right). As the training process progresses, the SADS increases, and the cross-lingual reasoning ability of the model improves. \textit{This indicates that the activation becomes more strongly related to the semantic meaning of inputs rather than linguistic information.} Consequently, the model can gradually better align inputs into the common semantic space of different languages.

\begin{figure}[!htb]
    \centering
    \includegraphics[width=0.49\textwidth]{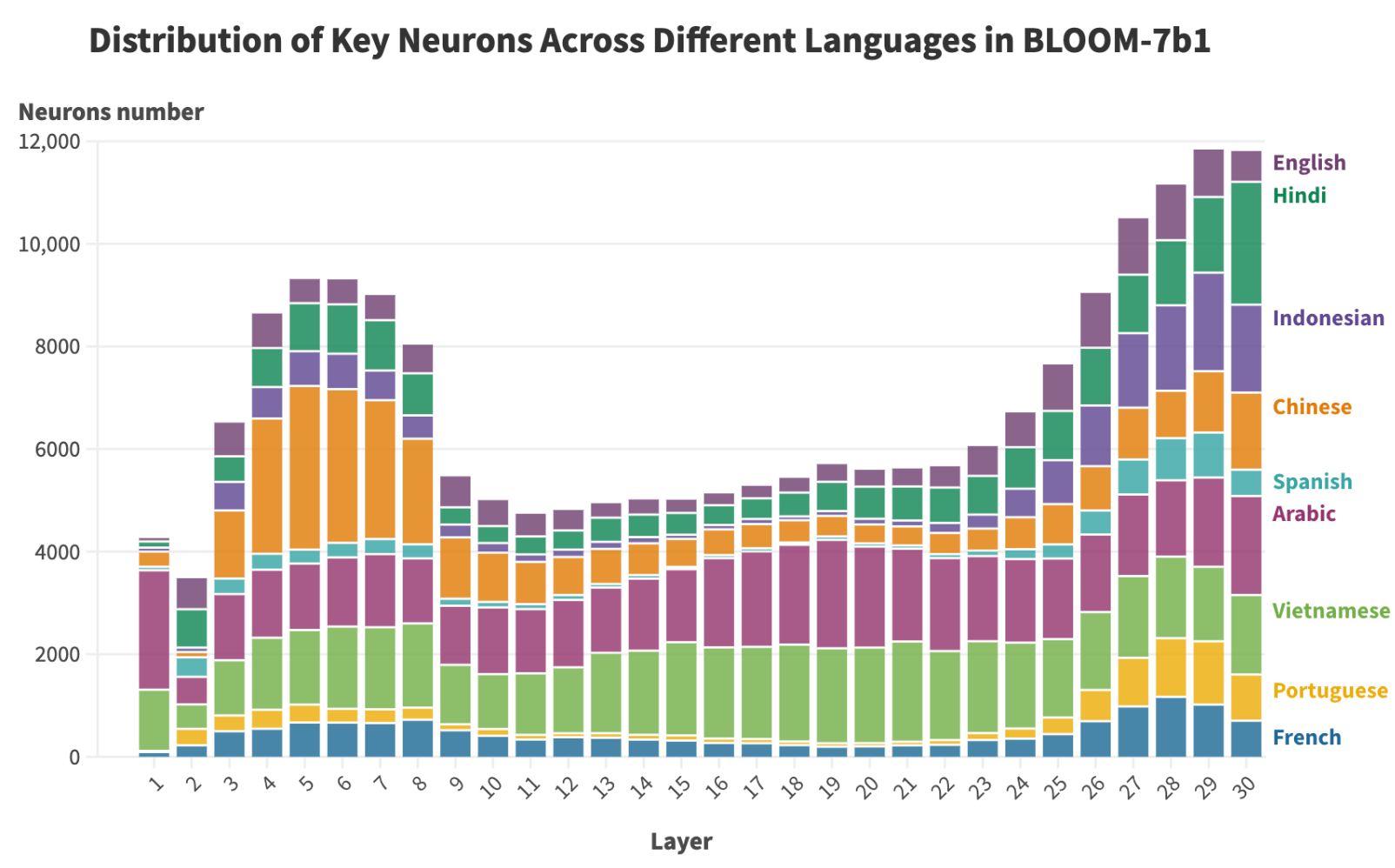}
    \caption{Distribution of Key Neurons Across Different Languages. }
    \label{ProportionLanguages}
    \vspace{-\baselineskip}
\end{figure}

By observing the size of the key linguistic region for each language (Figure \ref{ProportionLanguages}), we can see that languages with less similar linguistic features, such as Vietnamese, Arabic, and Chinese, which belong to very different language families and use distinct writing systems, have significantly larger key regions compared to their counterparts. This suggests that the model requires more effort to align these languages due to their distinct representations and linguistic features, making them generally more challenging to align within the mutual semantic space.

\subsection{Observation V: As the model scale grows, the model aligns languages better to the semantic space}

\begin{figure}[!htb]
    \centering
    \includegraphics[width=0.49\textwidth]{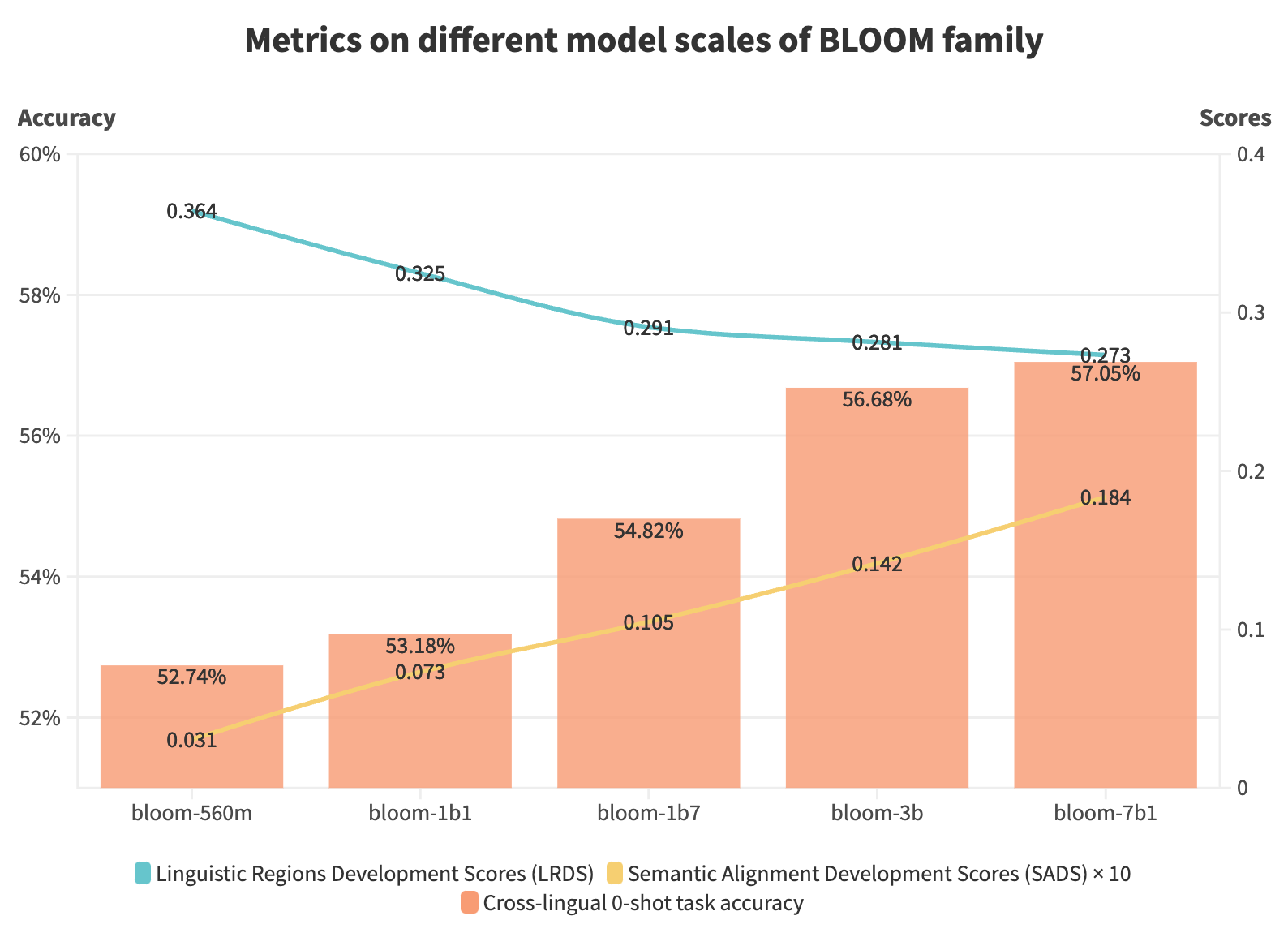}
    \caption{Metrics across different model scales of the BLOOM family. }
    \label{ModelScale}
    \vspace{-\baselineskip}
\end{figure}

We evaluated the LRDS, SADS, and cross-lingual zero-shot accuracy across different scales of BLOOM models (Figure \ref{ModelScale}). We can clearly see that as the model scale grows, the neurons' activation patterns become less related to language and more focused on the semantic meaning of inputs. At the same time, the cross-lingual reasoning ability of the model improves with larger scale. We conclude that as the model scale grows, LLMs can better align inputs in different languages to the common semantic space, enhancing their reasoning process within this shared semantic latent space.

\subsection{Extension to LLaMA2}

We performed the same experiments on LLaMA2. Since the intermediate checkpoints of LLaMA2 are not released, we conducted the analysis on different scales of the model, specifically on the 7b, 13b, and 70b models. We obtained similar results to the experiments performed on BLOOM, namely the existence of linguistic regions, semantic alignment, and the effects of scaling up the model. The results are shown in the Appendix \ref{llama2results}.

\section{Related Work}

\textbf{Multilingual Large Language Models. }Large language models (LLMs) such as GPT-4 \citep{openai2024gpt4}, LLaMA \citep{touvron2023llama}, and OPT \citep{zhang2022opt} have revolutionized natural language processing by demonstrating the ability to understand and generate nuanced text. Multilingual LLMs like BLOOM \citep{workshop2023bloom} and XLM-R \citep{conneau2020unsupervised} further extend these capabilities by learning universal representations from texts in multiple languages. These models typically use the transformer architecture \citep{vaswani2023attention}, incorporating multilingual data during pretraining to improve alignment and performance across languages \citep{qin2024multilingual}. To align different languages' capacity, various alignment strategies have been proposed, including alignment during pre-training \cite{blevins2022language, briakou2023searching, holmstrom-etal-2023-bridging}, supervised fine-tuning \cite{gao2024boosting, fu2022polyglot, cui2024efficient}, reinforcement learning from human feedback \cite{zeng2024tim, dong2023steerlm, sun2024salmon}, downstream fine-tuning \cite{aggarwal2024maple, rosenbaum-etal-2022-linguist, shaham2024multilingual}, prompt tuning \cite{qin2023crosslingual} and contrastive learning \cite{li-etal-2024-improving-cross-lingual-transfer}.

\textbf{Neuroscience. }In the field of neuroscience, considerable research has been conducted to understand how the human brain processes multiple languages. Studies have shown that polyglots—individuals who speak multiple languages—exhibit distinct patterns of brain activation for different languages \citep{10.1093/cercor/bhae049}. These findings suggest that different languages are stored and processed in different compartments of the brain \citep{PARADIS19851, article}. Additionally, research by \citet{XU2021104922} has indicated that similar brain activation patterns can occur when processing the same tasks in different languages, suggesting a common neural mechanism underlying multilingual processing.

\textbf{Linguistic and Semantic alignment. }Previous research indicates that the representations generated by multilingual encoder models are moderately language-agnostic \cite{pires-etal-2019-multilingual, libovicky-etal-2020-language}. Based on this assumption, \citet{yoon2024langbridge} introduced a LangBridge model to connect a multilingual encoder to a monolingual LLM, effectively achieving promising performance. They found equally that the efficacy of LangBridge stems from the language-agnostic characteristics of multilingual representations. \citet{ding2022simple} proposes targets to transfer English embeddings to virtual multilingual embeddings without semantic loss, thereby improving cross-lingual transferability. \citet{zhang2024unveiling} discovered a core region in LLMs that corresponds to linguistic competence, freezing the core linguistic region during further pre-training can mitigate the issue of catastrophic forgetting. \citet{wendler2024llamas} showcased that multilingual language models trained on unbalanced, English-dominated corpora use an abstract "concept space" laying closer to English as an internal pivot.

\section{Conclusion}
This paper explores the internal mechanisms of multilingual LLMs. We found that neuron activation patterns in LLMs are similar when processing the same language, allowing us to identify key neurons for specific languages. Deactivating these neurons significantly impairs performance in those languages. Additionally, we discovered that neuron activation patterns are similar when processing semantically identical sentences in different languages. This indicates that LLMs map these inputs into a common latent space. As training progresses, key linguistic regions become smaller and neuron activation becomes more focused on semantic meaning and less on language specifics. Key neurons are mainly located in the first and last layers, becoming denser in the first layers with training. Moreover, larger models align languages better, enhancing cross-lingual reasoning abilities. Our findings provide insights into the structural evolution of multilingual LLMs during training and scaling, offering a foundation for improving their cross-lingual capabilities.

\section*{Limitations}

While our study provides valuable insights into the internal mechanisms of multilingual LLMs, it has several limitations. Firstly, our analysis primarily focuses on the BLOOM and LLaMA2 models. Although these models are representative, the findings may not fully generalize to other multilingual LLM architectures. Future research should examine a broader range of models to validate our conclusions. In particular, investigating the evolution of the capabilities of monolingual models as they undergo continuous training with multilingual data could be a very interesting research subject. Secondly, we rely on specific datasets, such as the Bible dataset and XL-Sum, for our experiments. These datasets, while diverse, may not cover all linguistic nuances and complexities. Utilizing a wider array of datasets, including those with more diverse and low-resource languages, would provide a more comprehensive evaluation of model performance and neuron activation patterns. Thirdly, our methodology for identifying key neurons and measuring their contributions is based on averaged activation patterns and z-scores. This approach, while effective, may not capture all nuances of neuron interactions and their contributions to language processing. More sophisticated techniques, such as causal inference methods, could provide deeper insights into neuron functionality. Lastly, while we observed significant patterns related to semantic alignment and linguistic region efficiency, the underlying reasons for these patterns remain speculative. Further research is needed to establish causal relationships and to better understand the specific mechanisms through which LLMs achieve cross-lingual competence. These limitations highlight areas for future research to build upon our findings and enhance the understanding and capabilities of multilingual LLMs.

\section*{Acknowledgement}
This work is funded by the China NSFC Projects (92370206, 62120106006, U23B2057, and 62106142) and Shanghai Municipal Science and Technology Major Project (2021SHZDZX0102).

\bibliography{acl_latex}

\appendix

\section{Output examples after deactivating key linguistic region neurons}
In this section, we analyze the effects of deactivating specific languages (English, Chinese, and French) on BLOOM-7B1 (Table \ref{tab:appendix1}) and LLaMA2-7B (Table \ref{tab:appendix2}) models. For each language deactivation scenario, we input three separate samples (0., 1., 2.), and generate 64 tokens per sample. While the impact on English is minimal, likely due to its high-resource nature and the robust training models receive in English, the effects on Chinese and French are quite striking.

For Chinese, both BLOOM-7B1 and LLaMA2-7B fail to generate correct characters when the language is deactivated. The output consists of malformed or incomplete UTF-8 codes, indicating that the models are unable to construct valid Chinese text. In contrast, the impact on French, particularly in BLOOM-7B1, is even more interesting. When deactivated, the model produces a mixture of incorrect French and Spanish, revealing interference between the two languages. On LLaMA2-7B, deactivating French also corrupts the model’s ability to generate coherent text in the language.

This demonstrates that while high-resource languages like English maintain some level of robustness under deactivation, lower-resource or more specialized languages such as Chinese and French experience a much more pronounced degradation in their generation capabilities. This observation could have significant implications for multilingual model design and language-specific fine-tuning strategies.

\begin{table*}
    \centering
    \includegraphics[width=\textwidth]{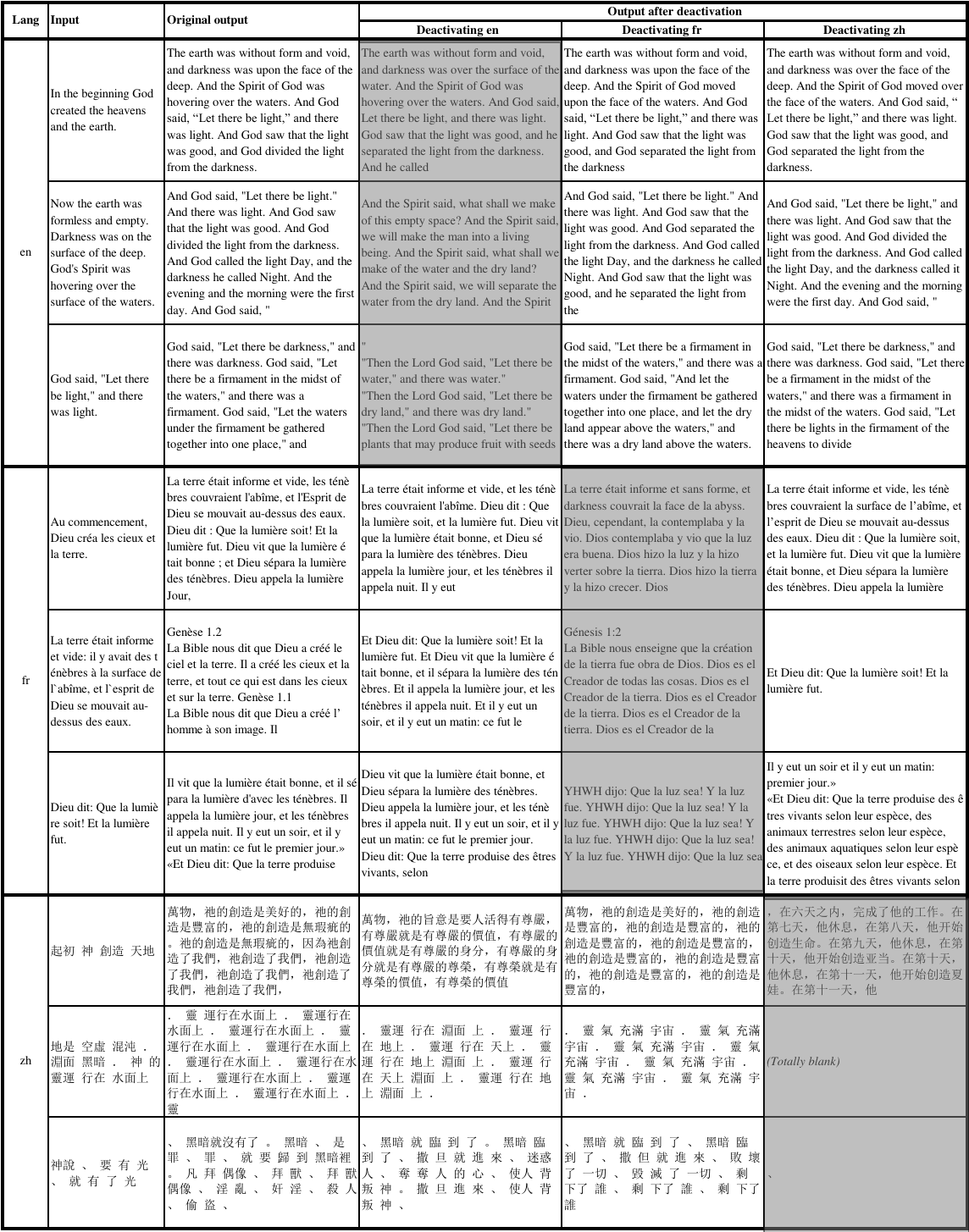}
    \caption{Examples of model output after deactivating the key linguistic neurons for English (en), French (fr), or Chinese (zh) on BLOOM-7B1.}
    \label{tab:appendix1}
\end{table*}

\begin{table*}
    \centering
    \includegraphics[width=\textwidth]{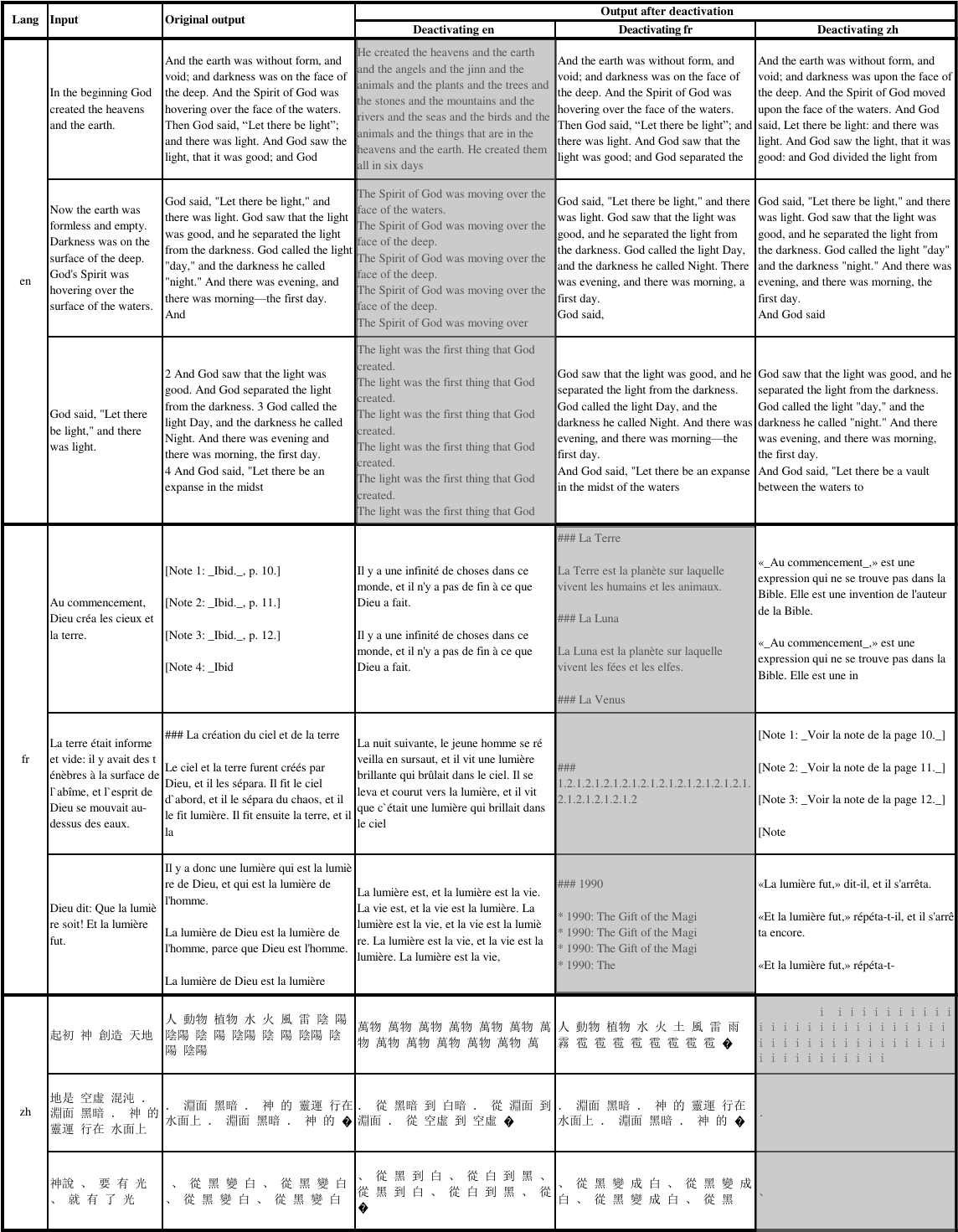}
    \caption{Examples of model output after deactivating the key linguistic neurons for English (en), French (fr), or Chinese (zh) on LLaMA2-7B.}
    \label{tab:appendix2}
\end{table*}

\section{Prompt format of XStoryCloze}
\label{sec:XStoryCloze}

We used a specific prompt to test the log likelihood of generating each option. If the log likelihood of generating the correct ending is higher, we infer that the model has correctly understood the story.

\textbf{Prompt:}

\texttt{[input\_sentence\_1, input\_sentence\_2, input\_sentence\_3, input\_sentence\_4, "The ending in English:" ]}

\textbf{Choice:}

\texttt{[Ending 1 in English, Ending 2 in English]}

We calculate the log likelihood of the model generating the two choice after the prompt.

\textbf{Example:}

\begin{enumerate}
\setlength{\itemsep}{0pt}
  \setlength{\parskip}{0pt}
    \item La madre le dijo a sus hijos que comerían en quince minutos. (Mother told her children it would be lunchtime in fifteen minutes.)
    \item Pero, entonces, recibió una llamada importante. (But then she got an important phone call.)
    \item Mientras hablaba, el perro llenó toda la cocina de barro. (While she was talking, the dog dragged mud all over the kitchen.)
    \item Los niños empezaron a fastidiar a su madre, que seguía al teléfono. (The kids started to pester their mother, who was still on the phone.)
\end{enumerate} 

\textbf{The ending in English:} 
\begin{enumerate}
\setlength{\itemsep}{0pt}
  \setlength{\parskip}{0pt}
    \item The mother felt quite frustrated.
    \item The children's behavior calmed the mother down.
\end{enumerate} 

\textbf{Correct Ending:} 1 

\textbf{Example Prompt:} La madre le dijo a sus hijos que comerían en quince minutos. Pero, entonces, recibió una llamada importante. Mientras hablaba, el perro llenó toda la cocina de barro. Los niños empezaron a fastidiar a su madre, que seguía al teléfono. The ending in English: 

\textbf{Choice 1 }(Target choice): The mother felt quite frustrated.

\textbf{Choice 2:} The children's behavior calmed the mother down.

The dataset supports evaluation on Arabic, Basque, Chinese, English, Hindi, Indonesian, Malay, Russian, Spanish, Swahili and Telugu. 

\section{Results of LLaMA2}
\label{llama2results}
\subsection{Similarity maps}

\begin{figure*}[!htb]
\resizebox{\textwidth}{!}{
\centering
	\begin{minipage}[t]{0.32\linewidth}
		\centering
		\includegraphics[width=\linewidth]{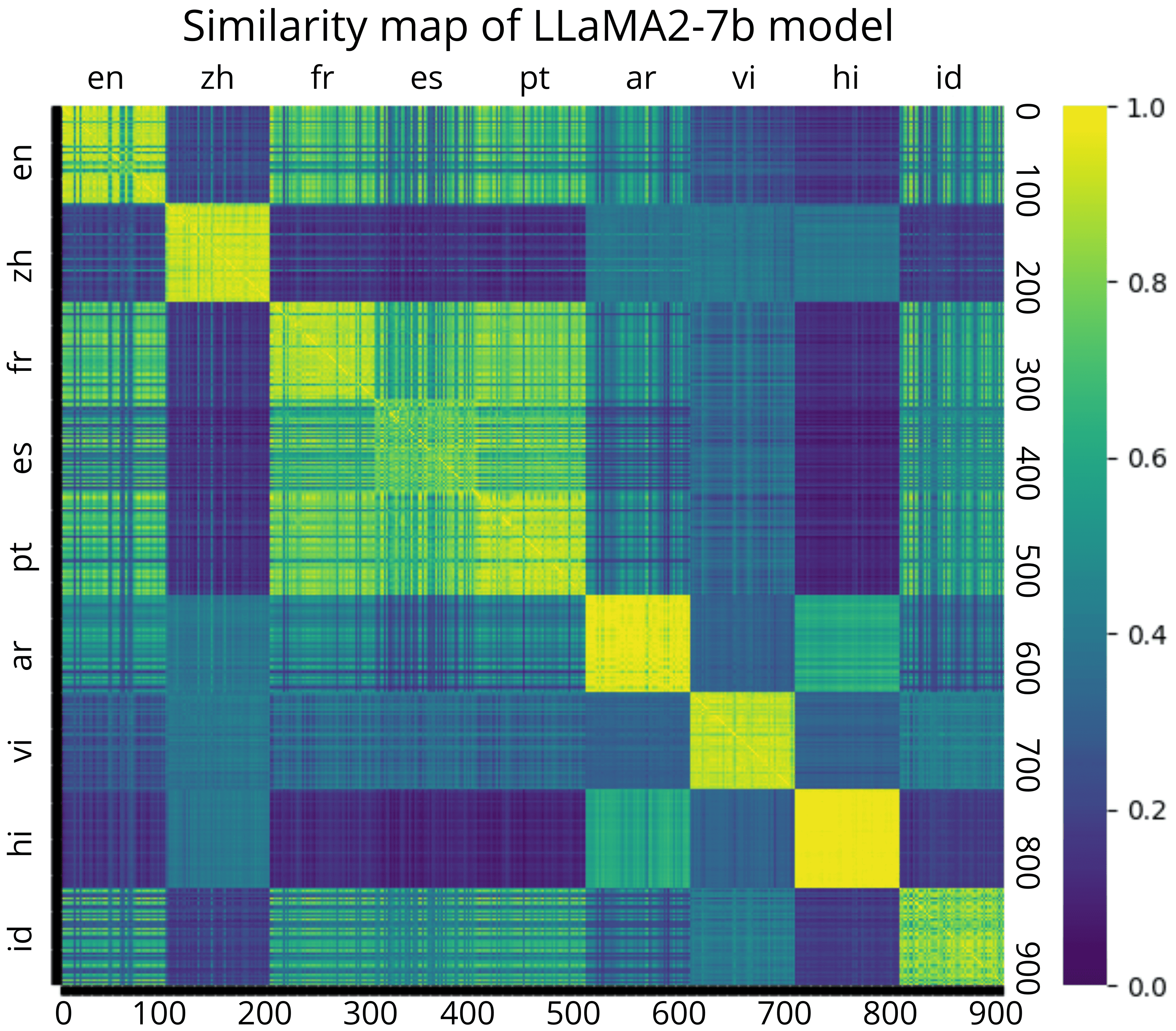}
	\end{minipage}
	\begin{minipage}[t]{0.32\linewidth}
		\centering
		\includegraphics[width=\linewidth]{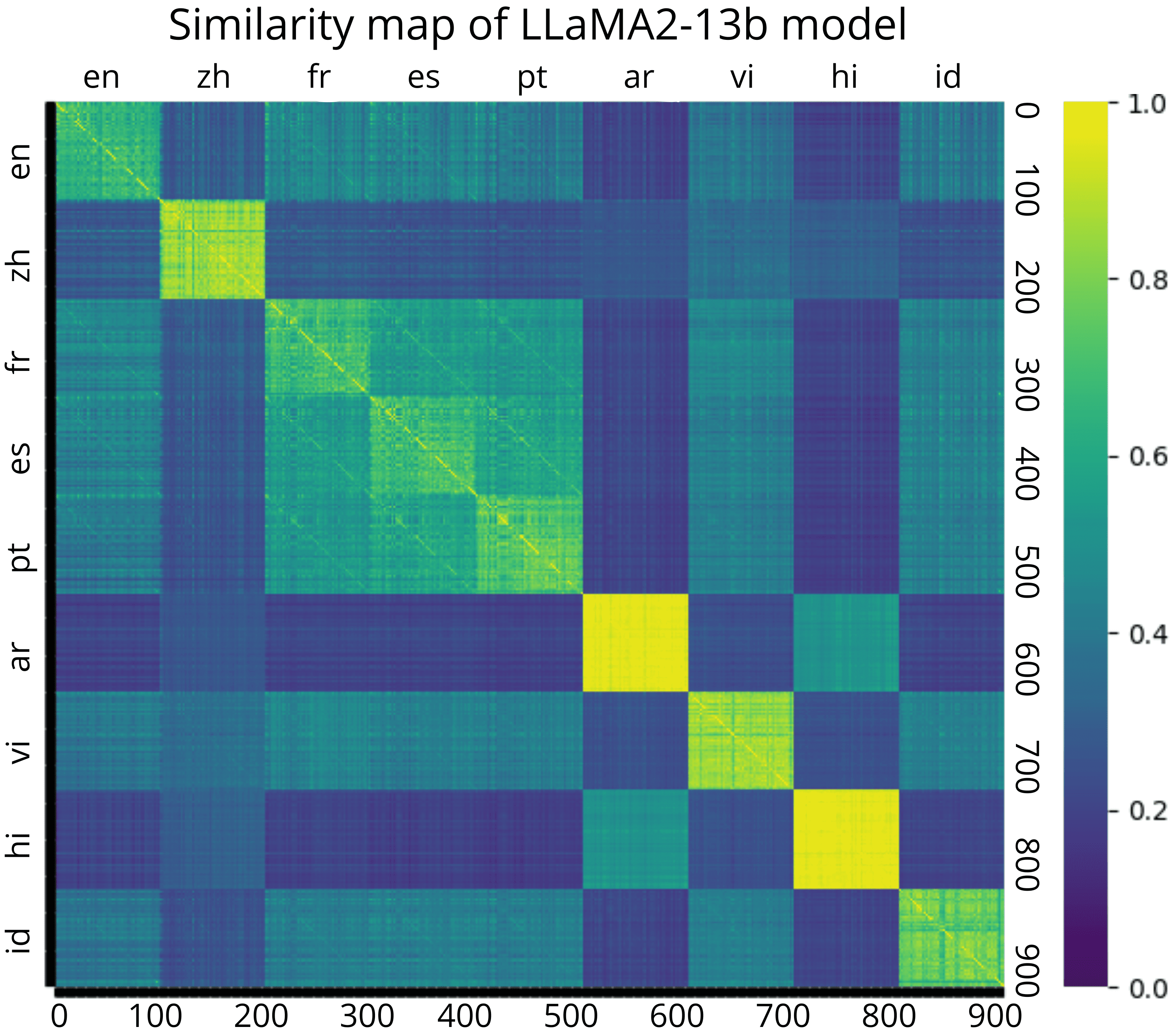}
	\end{minipage}
        \begin{minipage}[t]{0.32\linewidth}
		\centering
		\includegraphics[width=\linewidth]{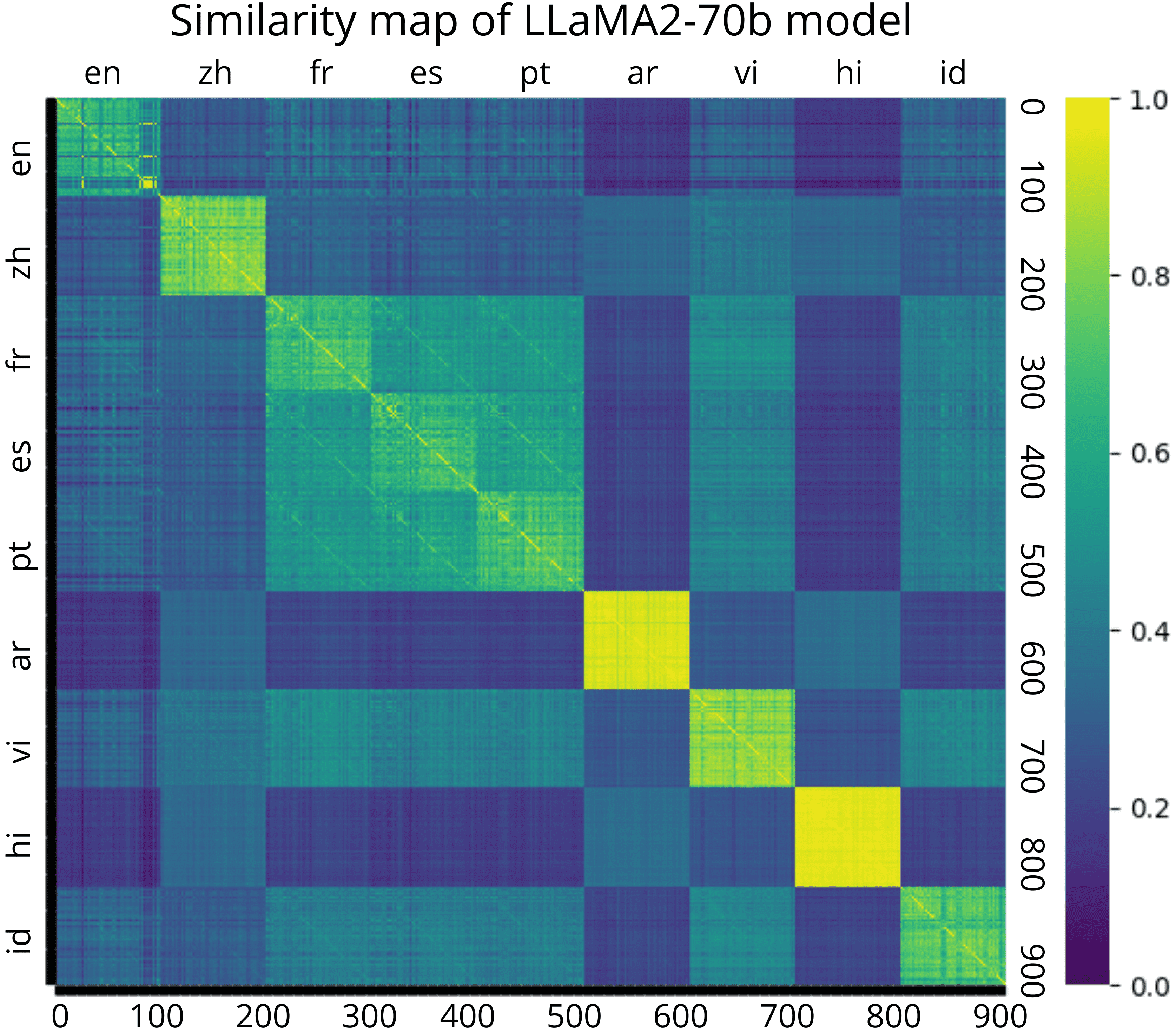}
	\end{minipage}}
	\caption{The similarity maps of LLaMA2 family. }
	\label{llama2similaritymap}
\end{figure*}

The similarity maps of the LLaMA2 family are presented in Figure \ref{llama2similaritymap}. We can clearly see the light blocks representing different languages and the light bands of semantically identical sentences. These bands become increasingly pronounced as the models scale up. However, they are not as prominent as in the BLOOM model family because LLaMA2 used only about 10\% non-English data during training.

\subsection{Layer-wise Semantic Alignment Development Scores (SADS) and Linguistic Regions Development Scores (LRDS)}

\begin{figure}[!htb]
    \centering
    \includegraphics[width=0.5\textwidth]{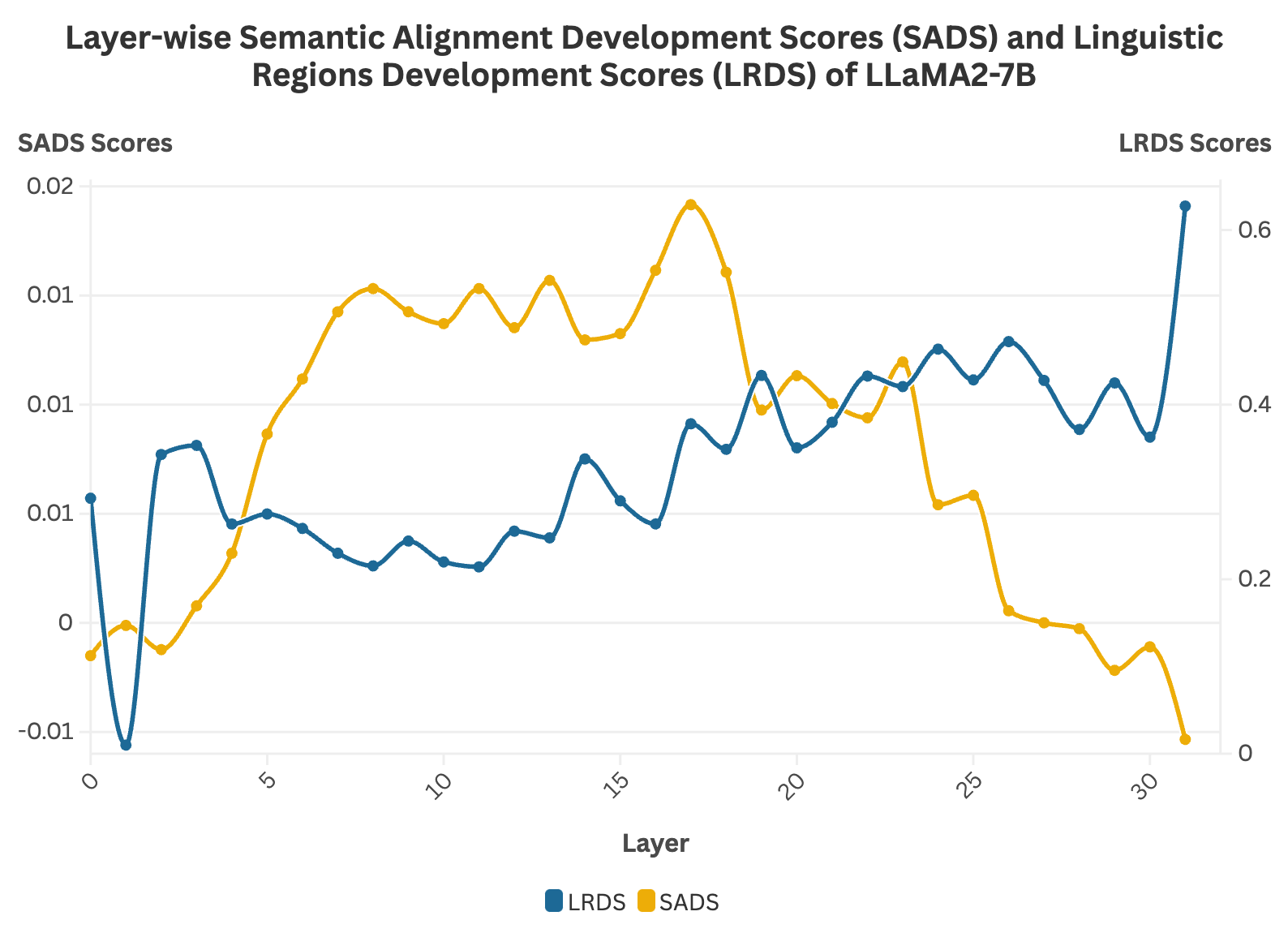}
    \caption{Layer-wise Semantic Alignment Development Scores (SADS) and Linguistic Regions Development Scores (LRDS) of LLaMA2-7B.}
    \label{layerwisellama}
\end{figure}

The results are presented in Figure \ref{layerwisellama}. The SADS scores follow a similar trend to those observed in the BLOOM-7b1 model. However, the LRDS scores exhibit more fluctuation: they are high in the final layers but relatively lower in the initial layer. This behavior may be attributed to the English-centric nature of the LLaMA2 training data.

\subsection{Key Linguistic Regions}

The percentage increase in perplexity after deactivating key linguistic region neurons for each language in the LLaMA2-7b model is shown in Table \ref{pplincreasellama7b}. As with the BLOOM models, we can see that there are key linguistic regions in LLaMA2-7b. When these regions are deactivated, the perplexity for the corresponding language increases significantly, while the perplexity for other languages remains largely unchanged.

\begin{table*}[!htb]
\resizebox{\textwidth}{!}{
\begin{tabular}{llllllllllll}
\rowcolor[HTML]{D9D9D9} 
\begin{tabular}[c]{@{}l@{}}Perplexity\\ increase \%↑\end{tabular}                    & \begin{tabular}[c]{@{}l@{}}Full model\\ Perplexity\end{tabular} & $\times$en                    & $\times$zh                    & $\times$fr                    & $\times$es                    & $\times$pt                    & $\times$ar                    & $\times$vi                     & $\times$hi                      & $\times$id                    & \begin{tabular}[c]{@{}l@{}}Random\\ 17\%\end{tabular} \\
en                                                                                   & 6.22                                                            & \cellcolor[HTML]{F8696B}10\%  & \cellcolor[HTML]{63BE7B}9\%   & \cellcolor[HTML]{FFEB84}3\%   & \cellcolor[HTML]{BCD780}7\%   & \cellcolor[HTML]{63BE7B}3\%   & \cellcolor[HTML]{63BE7B}6\%   & \cellcolor[HTML]{63BE7B}37\%   & \cellcolor[HTML]{63BE7B}15\%    & \cellcolor[HTML]{63BE7B}23\%  & \cellcolor[HTML]{F6E883}31\%                          \\
zh                                                                                   & 4.40                                                            & \cellcolor[HTML]{B8D67F}5\%   & \cellcolor[HTML]{F8696B}145\% & \cellcolor[HTML]{86C87D}3\%   & \cellcolor[HTML]{FFEB84}8\%   & \cellcolor[HTML]{83C77C}3\%   & \cellcolor[HTML]{FFEB84}8\%   & \cellcolor[HTML]{FFEB84}69\%   & \cellcolor[HTML]{FFEB84}23\%    & \cellcolor[HTML]{B5D57F}29\%  & \cellcolor[HTML]{C3D980}53\%                          \\
fr                                                                                   & 4.88                                                            & \cellcolor[HTML]{FED17F}7\%   & \cellcolor[HTML]{9FCF7E}10\%  & \cellcolor[HTML]{F8696B}46\%  & \cellcolor[HTML]{FFE283}13\%  & \cellcolor[HTML]{FFE984}5\%   & \cellcolor[HTML]{D2DE81}7\%   & \cellcolor[HTML]{C9DB80}55\%   & \cellcolor[HTML]{C5DA80}17\%    & \cellcolor[HTML]{B7D67F}29\%  & \cellcolor[HTML]{FFEB84}46\%                          \\
es                                                                                   & 5.66                                                            & \cellcolor[HTML]{FECC7E}7\%   & \cellcolor[HTML]{83C77C}9\%   & \cellcolor[HTML]{FFE984}4\%   & \cellcolor[HTML]{F8696B}67\%  & \cellcolor[HTML]{FFE583}9\%   & \cellcolor[HTML]{B7D67F}7\%   & \cellcolor[HTML]{D2DE81}56\%   & \cellcolor[HTML]{84C77C}16\%    & \cellcolor[HTML]{D4DE81}31\%  & \cellcolor[HTML]{FFEB84}48\%                          \\
pt                                                                                   & 5.55                                                            & \cellcolor[HTML]{FFDB81}6\%   & \cellcolor[HTML]{FFEB84}11\%  & \cellcolor[HTML]{FFE784}5\%   & \cellcolor[HTML]{FED380}19\%  & \cellcolor[HTML]{F8696B}101\% & \cellcolor[HTML]{FFEB84}7\%   & \cellcolor[HTML]{FFEB84}67\%   & \cellcolor[HTML]{E1E282}17\%    & \cellcolor[HTML]{FFEB84}36\%  & \cellcolor[HTML]{FFEB84}54\%                          \\
ar                                                                                   & 3.21                                                            & \cellcolor[HTML]{B4D57F}5\%   & \cellcolor[HTML]{FFE884}15\%  & \cellcolor[HTML]{FFEA84}4\%   & \cellcolor[HTML]{EFE683}8\%   & \cellcolor[HTML]{FFEB84}4\%   & \cellcolor[HTML]{F8696B}223\% & \cellcolor[HTML]{FFEB84}64\%   & \cellcolor[HTML]{FFEB84}57\%    & \cellcolor[HTML]{FFEA84}39\%  & \cellcolor[HTML]{D4DE81}62\%                          \\
vi                                                                                   & 3.34                                                            & \cellcolor[HTML]{BDD880}5\%   & \cellcolor[HTML]{FFE483}19\%  & \cellcolor[HTML]{63BE7B}3\%   & \cellcolor[HTML]{7BC47C}6\%   & \cellcolor[HTML]{7CC57C}3\%   & \cellcolor[HTML]{EEE683}7\%   & \cellcolor[HTML]{F8696B}1796\% & \cellcolor[HTML]{FFEB84}18\%    & \cellcolor[HTML]{FFEA84}38\%  & \cellcolor[HTML]{FFEB84}51\%                          \\
hi                                                                                   & 2.34                                                            & \cellcolor[HTML]{63BE7B}3\%   & \cellcolor[HTML]{FFDE82}26\%  & \cellcolor[HTML]{B2D47F}3\%   & \cellcolor[HTML]{63BE7B}6\%   & \cellcolor[HTML]{94CC7D}3\%   & \cellcolor[HTML]{FFE884}13\%  & \cellcolor[HTML]{A5D17E}48\%   & \cellcolor[HTML]{F8696B}27093\% & \cellcolor[HTML]{FFEB84}34\%  & \cellcolor[HTML]{63BE7B}54\%                          \\
id                                                                                   & 4.54                                                            & \cellcolor[HTML]{FFEB84}6\%   & \cellcolor[HTML]{94CC7D}9\%   & \cellcolor[HTML]{E7E482}3\%   & \cellcolor[HTML]{FFE383}12\%  & \cellcolor[HTML]{FFEA84}5\%   & \cellcolor[HTML]{FFEB84}8\%   & \cellcolor[HTML]{FFEA84}82\%   & \cellcolor[HTML]{FFEB84}22\%    & \cellcolor[HTML]{F8696B}517\% & \cellcolor[HTML]{F8696B}56\%                          \\
\cellcolor[HTML]{D9E1F4}\begin{tabular}[c]{@{}l@{}}Key Neuron \\ Number\end{tabular} & \cellcolor[HTML]{D9E1F4}                                        & \cellcolor[HTML]{F2F5FC}16817 & \cellcolor[HTML]{E3E9F7}23418 & \cellcolor[HTML]{FFFFFF}11343 & \cellcolor[HTML]{EAEFF9}20339 & \cellcolor[HTML]{FFFFFF}11007 & \cellcolor[HTML]{E9EEF9}20565 & \cellcolor[HTML]{91AADF}58145  & \cellcolor[HTML]{BBCAEB}40526   & \cellcolor[HTML]{A2B8E4}50881 & \cellcolor[HTML]{D9E1F4}59884                         \\
\rowcolor[HTML]{D9D9D9} 
\begin{tabular}[c]{@{}l@{}}Key Neuron \\ Percentage\end{tabular}                     &                                                                 & 4.8\%                         & 6.6\%                         & 3.2\%                         & 5.8\%                         & 3.1\%                         & 5.8\%                         & 16.5\%                         & 11.5\%                          & 14.4\%                        & 17.0\%                                               
\end{tabular}
}
\caption{Percentage increase in perplexity after deactivating key linguistic region neurons for each language in the LLaMA2-7b model. Each column corresponds to the deactivation of the key region for a specific language. The last column shows the results of deactivating a random 17\% (slightly higher than the maximum key neuron percentage, which is 16.5\% for \texttt{vi}) of neurons in the LLM.}

\end{table*}

\label{pplincreasellama7b}

\subsection{Comparison of Models of Different Scales}

\begin{figure}[!htb]
    \centering
    \includegraphics[width=0.49\textwidth]{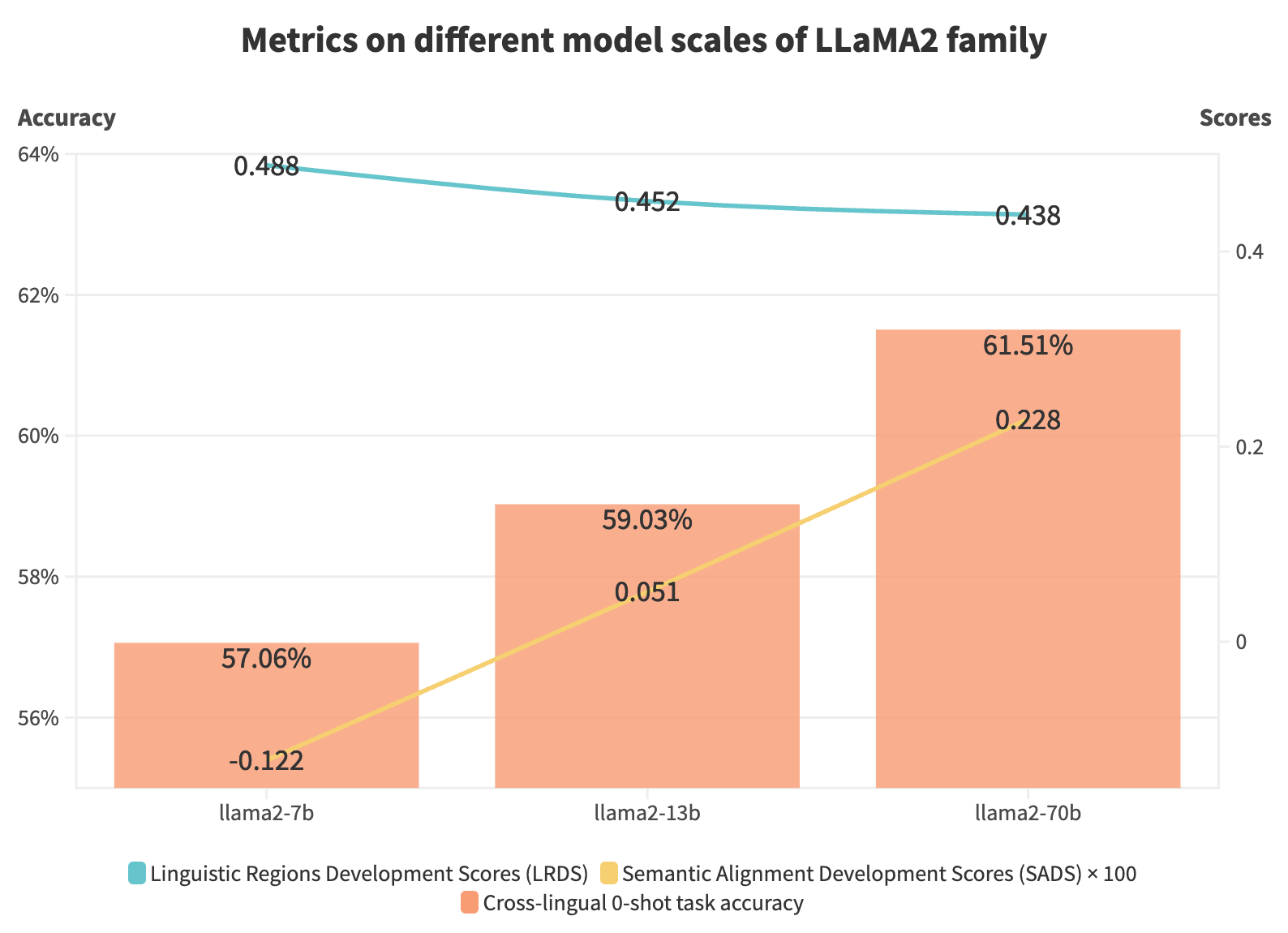}
\caption{Metrics across different model scales of the LLaMA2 family. \textbf{As the models scale up, activations become more language-agnostic and more semantically focused.}}
    \label{Model_scale_llama}
\end{figure}

We evaluate the results on LLaMA2 on different scale (Figure \ref{Model_scale_llama}). We may see that, as the models scale up, the activation patterns become more language-agnostic and more semantically focused, as in the BLOOM model family.

\begin{figure}[!htb]
    \centering
    \includegraphics[width=0.49\textwidth]{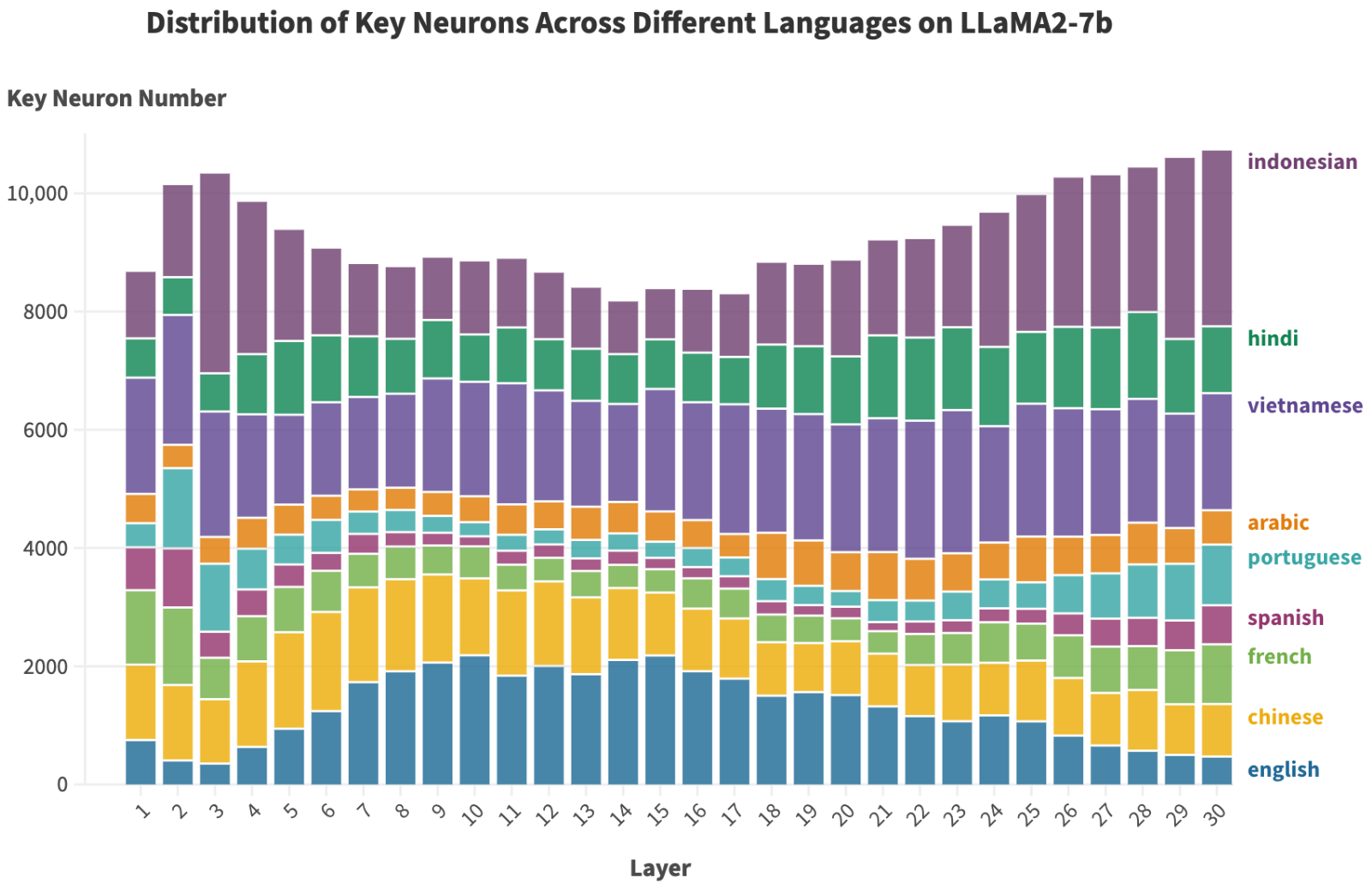}
    \caption{Distribution of Key Neurons Across Different Languages on LLaMA-7b.}
    \label{ProportionLanguages_llama7b}
\end{figure}

\end{document}